\documentclass[
  main, final
]{article}

     \PassOptionsToPackage{numbers, compress}{natbib}

\usepackage[main, final]{neurips_2026}

\usepackage[utf8]{inputenc} %
\usepackage[T1]{fontenc}    %
\usepackage{hyperref}       %
\usepackage{url}            %
\usepackage{booktabs}       %
\usepackage{amsfonts}       %
\usepackage{amsmath}        %
\usepackage{nicefrac}       %
\usepackage{microtype}      %
\usepackage{xcolor}         %
\usepackage{graphicx}       %

\usepackage[textsize=tiny
            ]{todonotes}
\usepackage{macros}
\usepackage{subcaption}
\usepackage{listings}
\title{Hierarchical Task Network Planning with LLM-Generated Heuristics}

\author{%
  Felipe Meneguzzi\\
  University of Aberdeen, UK\\
  PUCRS, Brazil
  \And
  Alexandre Buchweitz\\
  Independent Researcher\\
  Toronto, Canada
  \And
  Augusto B.\ Corrêa \\
  University of Oxford, UK
  \And
  Victor Scherer Putrich \\
  Saarland University, Germany
  \And
  André G. Pereira\\
  Universidade Federal do Rio Grande do Sul, Brazil
}

\begin{document}

\maketitle

\begin{abstract}
  \htn{} planning is a variation of classical planning where, instead of searching for a linear sequence of actions, an algorithm decomposes higher-level tasks using a method library until only executable actions remain.
  On the one hand, this allows one to introduce domain knowledge that can speed up the search for a solution through the method library.
  On the other hand, it creates challenges that go beyond those of classical state-space search.
  While recent research produced a number of heuristics and novel algorithms that speed up \htn{} planning, these heuristics are not yet as informative as those available in classical planning algorithms.
  We investigate whether large language models (LLMs) can generate effective search heuristics for \htn{} planning, extending the methodology of~\citeauthor{CorreaPereiraSeipp2025} \citeyear{CorreaPereiraSeipp2025} from classical to hierarchical planning.
  Using the \pytrich{} planner on seven standard total-order \htn{} benchmark domains, we evaluate heuristics generated by nine LLMs under domain-specific prompting and compare them against two established domain-independent heuristics in the same planner and against the \panda{} planner.
  A configuration fixed in advance solves more instances than either domain-independent heuristic, the best generated heuristics expand fewer nodes than both on most instances, and the virtual best over all models is on par with \panda{} in coverage.
\end{abstract}

\section{Introduction}
\label{sec:introduction}

Hierarchical Task Network (\htn{}) planning~\citep{ErolHendlerNau1994,GhallabNauTraverso2004a,BercherHollerBehnke2019} is a widely used formalism for structured, goal-directed tasks in robotics, game AI, and intelligent assistants.
Unlike classical planning, which searches over state transitions, \htn{} planning decomposes abstract tasks into sequences of primitive actions through \defterm{methods}, requiring search over a joint space of world states and task networks.
The decomposition hierarchy encodes domain knowledge directly, playing the role that explicit heuristics serve in classical planning: it constrains which actions the algorithm tries and in what order, without requiring a domain-independent estimate of goal distance.
This richer structure makes \htn{} planning substantially harder to scale without effective search guidance.

\htn{} planning is useful because many domains describe work as procedures rather than as a flat set of goal facts.
A robot mission, for example, may choose a survey, delivery, or repair procedure before expanding it into navigation, sensing, and manipulation steps; game agents and assistants use similar routines.
A classical encoding can represent the same primitive actions, but it often pushes procedural knowledge into extra predicates, control rules, or a large compiled model.
\htn{} models keep this knowledge explicit in tasks and methods, giving the LLM structured domain information that it can turn into search guidance.

Domain-independent heuristics offer guidance without requiring hand-crafted, domain-specific knowledge; the established methods include the admissible Task Decomposition Graph (\tdg{}) heuristic~\citep{BercherBehnkeHoellerEtAl2017}, the relaxed composition (RC) heuristics, which compile the \htn{} search state into a classical planning problem so that classical heuristics such as \textsc{FF} and \textsc{LMCut} apply~\citep{HoellerBercherBehnkeEtAl2019,HoellerBercherBehnkeEtAl2020}, delete- and ordering-relaxation heuristics~\citep{HoellerBercherBehnke2020}, and landmark-based counting (\lmcount{})~\citep{HoellerBercher2021,PutrichMeneguzziPereira2025}.
Despite these advances, heuristic search has only recently taken the lead in \htn{} planning.
\hypertension{}~\citep{MagnaguagnoMeneguzziDeSilva2025} won the first \htn{} track of the International Planning Competition (IPC) in 2020 with efficient recursive search and bookkeeping and no explicit heuristic function~\citep{BehnkeHoellerBercher2021}.
Heuristic progression planners were largely absent from that edition, so its result does not measure what a heuristic adds over good bookkeeping.
By the 2023 edition, planners built on the \panda{} heuristics won every total-order track~\citep{TaitlerAlfordEspasaEtAl2024}, which makes \panda{} the standard against which a new \htn{} heuristic must be measured.
Whether \htn{} heuristics can become as informative as their classical counterparts remains open.

Recent work shows that large language models (LLMs) can generate effective heuristic functions for classical planning as code~\citep{TuisovVernikShleyfman2025,CorreaPereiraSeipp2025}.
The methodology prompts an LLM with domain context, evaluates candidate heuristics on training problems, selects the best performer, and deploys it on unseen test problems.
The resulting heuristics outperform established domain-independent baselines and rival domain-specific methods, a striking result given the absence of any explicit heuristic engineering.
Extending this methodology to \htn{} planning poses additional challenges.
A heuristic for \htn{} search must reason not only over the world state but also over the current \defterm{task network}: the set of pending tasks, their ordering constraints, and the applicable decomposition methods.
An LLM must therefore grasp hierarchical decomposition semantics to produce useful guidance, a harder task than reasoning over state-only features.
Thus, this paper adapts the LLM heuristic generation methodology of~\citet{CorreaPereiraSeipp2025} to total-order \htn{} planning.
We keep the generate-evaluate-select pipeline deliberately close to theirs, so that differences in outcome are attributable to the hierarchical setting rather than to changes in the pipeline, and we make the following contributions.
\begin{enumerate}
  \item We implement the pipeline within the \pytrich{} planner~\citep{PutrichMeneguzziPereira2025} with a heuristic interface that exposes the task network alongside the world state, and a prompt that teaches the LLM to use it (Section~\ref{sec:method}).
  \item We evaluate heuristics generated by nine LLMs on seven benchmark domains against domain-independent heuristics within the same planner and against \panda{} (Section~\ref{sec:setup}).
  A configuration fixed in advance solves more problems than the domain-independent heuristics in \pytrich{}, and the virtual best is on par with \panda{} (Section~\ref{sec:results}).
  \item We show that the generated heuristics expand fewer nodes than those heuristics on most shared instances, that their margin shrinks as the decomposition structure of a domain becomes more regular, that no single model dominates, and that greedy search exploits them best (Sections~\ref{sec:results} and~\ref{sec:discussion}).
\end{enumerate}

\section{HTN Planning}
\label{sec:htn}

An \htn{} planning problem is a tuple $\htnproblem = \tuple{s_0, \tasknetwork_I, F, A, C, \methods}$, where $F$ is a set of ground fluents, $A$ is a set of primitive actions with preconditions and effects over $F$, $C$ is a set of compound tasks, $\methods$ is a set of methods, $\tasknetwork_I$ is the initial task network, and $s_0 \subseteq F$ is the initial state~\citep{ErolHendlerNau1994,GhallabNauTraverso2004a,PutrichMeneguzziPereira2025}.
A \defterm{task network} $\tasknetwork = \tuple{T, \prec, \alpha}$ consists of a finite set of task identifiers $T$, a strict partial order $\prec$ over $T$ encoding ordering constraints, and a labelling function $\alpha : T \to A \cup C$ that assigns each identifier to a primitive action or compound task.
A \defterm{method} $\method = \tuple{c, \tasknetwork_m} \in \methods$ decomposes a compound task $c \in C$ into a subnetwork $\tasknetwork_m$; applying $\method$ replaces the occurrence of $c$ in the current network with $\tasknetwork_m$.
A \defterm{solution} is a sequence of primitive actions obtained by repeatedly selecting and decomposing compound tasks until the network contains only primitive tasks, such that the resulting sequence is executable from $s_0$.

This paper focuses on \defterm{total-order} \htn{} planning, in which every task network imposes a strict linear order on its tasks.
Total-order \htn{} is widely studied~\citep{BercherHollerBehnke2019} and is the main setting used in the main IPC HTN planning track~\citep{BehnkeHoellerBercher2021,TaitlerAlfordEspasaEtAl2024}, which supplies the domains and problems for our evaluation.
Our evaluation uses problems encoded in \hddl{}~\citep{HoellerBehnkeBercherEtAl2020}, the standard input language for \htn{} planners.
We use the \pytrich{} planner~\citep{PutrichMeneguzziPereira2025} as our experimental platform because it exposes a clean Python interface for plugging in custom heuristic functions, making it straightforward to deploy LLM-generated code.

The \tdg{} heuristic~\citep{BercherBehnkeHoellerEtAl2017} estimates the cost to solve the current task network by computing a relaxed reachability bound over the Task Decomposition Graph, a precomputed structure that encodes which primitive actions can arise from each compound task.
It ignores delete effects and ordering constraints, yielding an admissible but often weak estimate.

Landmark-based counting (\lmcount{})~\citep{HoellerBercher2021} extracts \defterm{landmarks}, facts or tasks that must occur on every solution path, from the task network structure.
The heuristic value is the count of landmarks not yet achieved.
\citet{PutrichMeneguzziPereira2025} extend this approach with bidirectional landmark generation, which finds additional landmarks at polynomial cost.

Both heuristics are domain-independent and exploit only the syntactic structure of the \hddl{} encoding, requiring no user-supplied domain knowledge.
They serve as the primary algorithmic baselines in our evaluation.
To follow the protocol of \citeauthor{CorreaPereiraSeipp2025} of measuring their heuristics against the strongest available planners, we compare against the progression-based \panda{} planner~\citep{HoellerBercherBehnkeEtAl2020}, the state of the art in \htn{} planning.
\panda{}'s relaxed composition (RC) heuristics compile the current search state into a classical planning problem and estimate its cost with a classical heuristic.

\section{LLM-Generated Heuristics for HTN Planning}
\label{sec:method}

For each domain, an LLM writes a Python class with two methods: a preprocessing method that runs once before search, and an evaluation method that maps a search node, a state together with its remaining task network, to a non-negative estimate of the cost to complete it.
This class is the heuristic, and the planner uses it as it would any hand-written one.
We obtain these classes by adapting to \htn{} planning the pipeline of \citet{CorreaPereiraSeipp2025}, who show that LLMs can serve as heuristic engineers for classical planning.
Their methodology asks an LLM to write a Python function that maps a planning state to a numeric estimate of the distance to the goal.
They generate a batch of candidate functions by prompting the model with the domain encoding and a description of the heuristic interface, then discard candidates that raise exceptions or time out.
Next, they evaluate survivors on a set of training problems, and select the one that minimises node expansions for deployment on unseen test problems.
The pipeline is model-agnostic, and the selection step retains the most effective candidate whichever LLM wrote it.

Adapting this methodology to \htn{} planning requires two changes.
First, the heuristic must accept a task network in addition to the world state, since the remaining work comprises the pending tasks and their decomposition structure rather than a goal condition over fluents alone.
Second, the LLM must reason about the hierarchical semantics of the domain, namely which compound tasks remain, how they may be decomposed, and whether the current ordering of tasks suggests a long or short path to completion.
These demands are qualitatively harder than reasoning over flat state features, making \htn{} planning a more challenging test for LLM-generated heuristics.

\subsection{Heuristic Function Interface}
\label{sec:interface}

\pytrich{} exposes a Python \texttt{Heuristic} base class with two required methods.
The initialization method is called once before search and is used for preprocessing; the evaluation method is called at every expansion and must return a non-negative estimate of remaining solution cost.
The split keeps per-node evaluation cheap, since expensive operations run once in initialization.
The cost a heuristic estimates is the number of primitive actions the planner still needs to complete the task network, since \pytrich{} adds one unit for each primitive action it applies and nothing for a decomposition.
\pytrich{} never prunes a node based on its heuristic value, infinity included, so a candidate that overestimates or returns infinity slows the search but cannot make it incomplete.

The interface also exposes both parts of an \htn{} search state.
From \texttt{model}, the heuristic receives the grounded structure (facts, primitive operators, abstract tasks, decomposition methods, and goals).
From \texttt{node}, it receives the current state and the remaining task network.
In particular, states and operator conditions are represented as bitsets, which enables constant-time fact checks and efficient relaxed computations.

The most important distinction from the classical planning interface of~\citet{CorreaPereiraSeipp2025} is explicit access to the \htn{}-specific task network, that is, the pending abstract tasks and the decomposition methods applicable to each.
A heuristic that ignores the task network degenerates to a classical-planning heuristic and loses all \htn{}-specific guidance; exploiting this structure is the central challenge the LLM must address to produce useful estimates.
We include a detailed explanation of the API in Appendix~\ref{app:api} for reproducibility.

\subsection{Prompt Design}
\label{sec:prompts}

We use a domain-specific prompt regime in which the LLM receives the domain name, the full domain \hddl{} file, the smallest and largest benchmark instances of the domain, and a worked example of a heuristic for a different domain.
Following \citet{CorreaPereiraSeipp2025}, the example teaches the interface and not the domain, and Appendix~\ref{app:ablation} shows that it improves code validity without changing search performance.
\begin{figure}[t]
  \begin{minipage}[t]{0.40\textwidth}
\begin{lstlisting}[style=promptmd]
You are an expert in hierarchical planning and heuristic design. Your task is to analyze the provided domain and create an effective domain-dependent heuristic for the **Depots** domain.
[...]
## 12. Your Task: Required Design Procedure
Follow these steps and **write the justification in top-of-file comments** of your generated heuristic:
1. **Name the bottleneck.** Read the Depots HDDL and identify the single constraint that most serializes work. One-sentence comment.
2. **List 2-3 independent admissible lower bounds.** For each, write a one-line admissibility sketch: why is this term a lower bound on unavoidable cost?
3. **Implement `initialize(self, model)`.** Parse fact names, build per-predicate id maps, precompute goal structures [...] Preprocessing here is free.
4. **Implement `__call__(self, node)` to return `max(term1, term2, term3)`.** Each term must use O(1) bit tests. No iteration over model-sized collections.
5. **Add tie-breaking penalties** [...] Use sub-unit magnitudes (`* 0.001`) if preserving admissibility matters [...]
\end{lstlisting}
  \end{minipage}\hfill
  \begin{minipage}[t]{0.58\textwidth}
\begin{lstlisting}[style=genpython]
class DepotsAnthropicOpus46Heuristic9(Heuristic):
    [...]
    def _estimate_put_on_cost(self, crate, target, state):
        key = (crate, target)
        if key in self.on_fact_id and \
                (state >> self.on_fact_id[key]) & 1:
            return 0  # Already on target - nop
        cost = 0
        if not self._is_clear(target, state):
            blocking = self._count_stack_above(target, state)
            cost += blocking * 4  # Lift+Load+Unload+Drop
        target_place = self._get_surface_place(target)
        crate_place = self._get_crate_location(crate, state)
        if crate_place is not None:
            if not self._is_clear(crate, state):
                above = self._count_stack_above(crate, state)
                cost += above * 4
            cost += 2 if crate_place == target_place else 5
        elif self._is_crate_in_truck(crate, state):
            cost += 3  # Drive + Unload + Drop
        [...]
        return cost

    def _compute(self, state, task_network):
        [...]
        for task in task_network:
            [...]
            if task_name == "do_put_on":
                task_cost += self._estimate_put_on_cost(
                    args[0], args[1], state)
        [...]
        return max(goal_cost, task_cost,
                   unsatisfied_goals + clearing_cost)
\end{lstlisting}
  \end{minipage}
  \caption{Left, the preamble and the design procedure of the prompt, abridged, as sent for Depots.
           Right, the core of one heuristic Claude Opus~4.6 wrote from it, abridged.
           The cost of a pending \texttt{do\_put\_on} task depends on where the crate is and on the stack above its target.
           We denote elisions as \texttt{[...]}; Appendix~\ref{app:heuristics} lists the heuristic at greater length beside a failed one.}
  \label{fig:prompt-excerpt}
\end{figure}

The prompt directs the LLM through four steps: identify the planning bottleneck by determining which compound tasks decompose into many primitives; list two or three independent lower bounds on the remaining cost and combine them by taking their maximum, or their sum when they count disjoint work; implement the result using the two-method interface, keeping the cost per-node low by moving all loops to the preprocessing step; and add tie-breaking penalties for symmetric states, keeping them below one unit when the estimate must remain a lower bound.
Figure~\ref{fig:prompt-excerpt} reproduces this part of the prompt beside the core of a heuristic Claude Opus~4.6 wrote from it for Depots, which estimates the cost of each pending task from the state rather than from its cheapest method.

Several technical details require explicit treatment in the prompt because LLMs otherwise default to incorrect conventions when working with \pytrich{}.
The format of grounded fact names differs from \hddl{} predicate syntax, and the prompt illustrates this difference with wrong-versus-right examples.
The prompt also specifies how to query goal facts via the bitwise state encoding, provides correct import paths for the heuristic base class and model types, and gives full method signatures with type annotations.
The prompt asks the LLM to return its response in a structured, parseable format so that candidates can be saved without additional text processing.

The prompt can also carry a per-domain hint block with three parts: representation caveats introduced by grounding; the dominant search bottleneck of the domain; and guidance on lower bounds, symmetry-breaking terms and penalty scales.
We wrote the blocks from the records of our development sessions, in which the model summarised the failure modes it met (from which we kept the summary as hint text), so they play the part of the handwritten checklist of \citet{CorreaPereiraSeipp2025}.
We treat the hints as an ablation condition rather than as part of the pipeline.
The results of Section~\ref{sec:results} use the prompt without hints, and Appendix~\ref{app:ablation} shows their effect with Gemini~3 Flash.
In summary, they leave coverage and code validity unchanged and reduce expanded nodes on about three quarters of the shared instances, mostly in Barman and Rover.
Appendix~\ref{app:hints} reproduces the blocks.

We also implement an iterative refinement variant that augments the base prompt with the previous candidate code, its empirical results against the \tdg{} baseline (expanded nodes, wall time, solution length, and exit status), and auto-generated guidance keyed to the observed failure mode.
A candidate that times out receives advice to move all computation to the preprocessing step; a candidate that performs worse than \tdg{} receives advice to add state-awareness or to aggregate decomposition costs across the full task network.
Appendix~\ref{app:prompts} has the full prompt text and the refinement template.

\subsection{Heuristic Selection}
\label{sec:selection}

For each domain and each LLM model, we generate a batch of candidate heuristics by independent one-shot invocations of the domain-specific prompt, with no feedback between them.
The batch holds twenty candidates for Claude Opus~4.6, six for Gemini 3 Flash on the six IPC 2020 domains, and between four and forty for the other models.
Appendix~\ref{app:expdetails} gives the counts for every model and domain.
Each candidate runs on the smallest benchmark problem for the domain under a 60-second wall-clock timeout.
We discard candidates that fail to parse, raise runtime exceptions, exceed memory limits, or time out.
Among the survivors, we select the candidate that minimises expanded nodes on the training problem, breaking ties in favour of the shorter solution.
Selection on one instance filters out candidates that crash or misuse the interface, but it does not rank the survivors, several of which tie on the selection instance in Robot and Towers. 

We then evaluate the selected candidate without modification on all benchmark problems for that domain, including the selection instance itself.
The selection criterion is model-agnostic, since it requires only that the heuristic run without error and produce a result on the single training instance, and it does not reward interpretability.
This design means that a weaker model contributing just one viable heuristic can still improve the virtual best over all models, and the selection step acts as an automatic quality filter without requiring the experimenter to implement heuristic code.

We leave a systematic evaluation of the iterative refinement variant to future work; all quantitative results in Section~\ref{sec:results} use heuristics generated by the base one-shot prompt.

\section{Experimental Setup}
\label{sec:setup}

We briefly describe the main points of our experimental setup next.\footnote{Code, generated heuristics, result files and scripts: \url{https://doi.org/10.5281/zenodo.23931424}}
Appendix~\ref{app:expdetails} contains more details, such as specific LLM versions, parameters, and resources used.

\paragraph{Domains.}
We evaluate on seven total-order \htn{} domains encoded in \hddl{}~\citep{HoellerBehnkeBercherEtAl2020}.
Six of them, Barman, Blocksworld, Depots, Robot, Rover and Towers, come from the IPC 2020 \htn{} track~\citep{BehnkeHoellerBercher2021} and remain in the 2023 track~\citep{TaitlerAlfordEspasaEtAl2024}.
The seventh, Lamps, is one of the two total-order domains that the 2023 track added.
We evaluate every benchmark instance of each domain, 180 problems over the seven domains, and Table~\ref{tab:coverage} gives the number per domain.
The Gemini API no longer serves Gemini 2.0 Flash and Gemini 3 Pro, so Lamps has candidates from Claude Opus~4.6, Gemini 2.5 Pro and Gemini 3 Flash only.

\paragraph{Models.}
We evaluate nine LLMs spanning three providers: Claude Opus~4.6, Claude Sonnet~4.5, Gemini 2.0 Flash, Gemini 2.5 Pro, Gemini 3 Flash, Gemini 3 Pro, GPT-4o, GPT-5, and GPT-5.2.
For each domain-model pair, we generate heuristic candidates using provider-default sampling and evaluate them under the protocol of Section~\ref{sec:selection}.
A list of models, generation settings, and model-specific failure accounting are reported in Appendix~\ref{app:expdetails}; summary failure analysis appears in Section~\ref{sec:models_results}.
We compare the LLM heuristics against two domain-independent heuristics in \pytrich{} and four configurations of \panda{}.

\paragraph{Baseline Planners and Heuristics.}
The \tdg{} heuristic~\citep{BercherBehnkeHoellerEtAl2017}, as implemented in \pytrich{}, precomputes for every compound task the cheapest decomposition into primitive actions over the task decomposition graph, weighting each primitive action by its cost and each method application by one unit, and estimates a node by the sum of these costs over the tasks remaining in its task network.
\lmcount{}, also in \pytrich{}, counts the landmarks not yet achieved and runs with the bidirectional landmark generation of \citet{PutrichMeneguzziPereira2025}.
\panda{}~\rcff{}, \rclmcut{} and \rcadd{}~\citep{HoellerBercherBehnkeEtAl2020} are the relaxed composition (RC) heuristics of \panda{}, which compile the current search state into a classical planning problem and estimate it with the FF~\citep{HoffmannNebel2001}, LMCut~\citep{HelmertDomshlak2009} and additive heuristics in turn.
\pandalmcount{} is \panda{}'s own landmark-count heuristic with bidirectional landmarks.
Both planners use one unit of path cost per primitive action and nothing per decomposition.
The LLM heuristics are Python classes written against the \pytrich{} interface, and \panda{} offers no hook for them, so the comparison across planners rests on coverage under the same limits.
We apply the same virtual-best treatment to the \panda{} configurations as to the LLM heuristics.

\paragraph{Search Algorithms.}

We evaluate each LLM-generated heuristic and each \pytrich{} baseline under three search algorithms: Standard A* ($f = g + h$), Greedy Best-First Search (GBFS, $f = h$ only, favouring nodes with the lowest heuristic value), and Weighted A* with weight $w = 5$ (WA*, $f = g + 5 \cdot h$), which trades plan optimality for faster search.
The weight is the same for every system, LLM heuristics and baselines alike, and we do not tune it.
For each system we report the \emph{virtual best} over these three algorithms: a problem counts as solved if any algorithm finds a solution, and we record the node count from the algorithm that solves it with the fewest expansions.
This reflects a realistic deployment scenario in which one can choose the search algorithm freely.
Results disaggregated by algorithm appear in Section~\ref{sec:results}.

\paragraph{Evaluation Metrics and Resources.}

We report three metrics.
\textit{Coverage} is the number of benchmark problems we solve within the time and memory limits; it is the primary metric, reported per domain and in aggregate.
\textit{Node expansions} is the number of search nodes expanded before finding a solution; lower is better, as it measures heuristic guidance quality independently of hardware speed.
We report the median over solved problems and head-to-head win counts on the common set of problems solved by both compared systems.
\textit{Plan length} is the number of primitive actions in the returned solution; we report head-to-head comparisons on the common solved set.
We run our experiments with 8~GB RAM and a 30-minute runtime limit per problem. All runs use 1 CPU core.
We run \panda{} experiments under identical resource constraints using a single build of \texttt{pandaPIengine}; additional execution details are provided in Appendix~\ref{app:expdetails}.

\section{Results}
\label{sec:results}

\begin{figure}[t]
    \centering
    \begin{subfigure}[t]{0.48\textwidth}
        \centering
        \includegraphics[width=\linewidth]{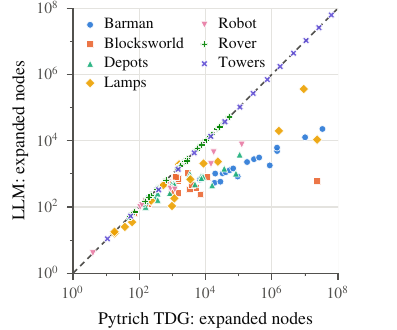}
        \caption{Expanded nodes for \llmvb{} vs.\ \pytrich{}~\tdg{}.}
        \label{fig:scatter_expanded_nodes_tdg}
    \end{subfigure}
    \hfill
    \begin{subfigure}[t]{0.48\textwidth}
        \centering
        \includegraphics[width=\linewidth]{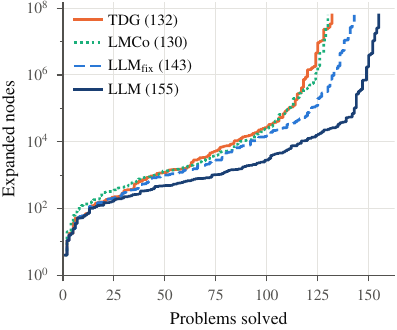}
        \caption{Problems solved within a given node-expansion budget in \pytrich{}.}
        \label{fig:main-pytrich-cactus}
    \end{subfigure}
    \par\medskip
    \begin{subfigure}[t]{\textwidth}
        \centering
        \includegraphics[width=\linewidth]{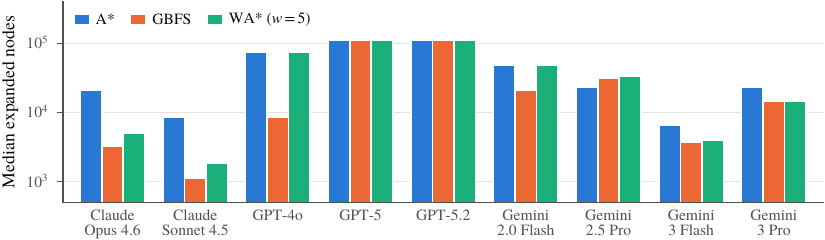}
        \caption{Median expanded nodes (log scale) by search algorithm for each LLM model.}
        \label{fig:algorithm_comparison}
    \end{subfigure}
    \caption{Search efficiency and algorithm effects: per-problem expanded nodes of \llmvb{} against \pytrich{}~\tdg{} (a), cumulative coverage of \tdg{}, \lmcount{}, \llmfixed{} and \llmvb{} against node-expansion budget (b), and median expanded nodes by search algorithm for each LLM model (c).
             In (a), points below the diagonal indicate an \llmvb{} advantage.}
    \label{fig:main_results}
\end{figure}

\subsection{Coverage}
\label{sec:coverage}

Table~\ref{tab:coverage} reports per-domain and aggregate coverage for all systems over the seven domains.
Every column but one is a virtual best over the three search algorithms (GBFS, A$^*$, WA$^*$), so a problem counts as solved if any algorithm solves it.
The \panda{} block shows the three relaxed-composition engines, \rcff{}, \rclmcut{} and \rcadd{}, the last with the additive estimate, together with \pandalmcount{}, \panda{}'s own landmark-count heuristic.
The \pytrich{} block shows \tdg{}, \lmcount{}, which is bidirectional landmark counting~\citep{PutrichMeneguzziPereira2025}, the best single model, Claude Opus~4.6 (\coiv{}), and two columns for the LLM heuristics as a whole.
\llmvb{} is the virtual best over the five main models and all their candidate heuristics.
This is an upper bound that no deployed system can reach, since it chooses the model after seeing the result.
\llmfixed{} is its deployable counterpart with one model and one search algorithm per domain, chosen on the smallest instance before running any other instance, taking the best of that model's candidates per instance.
We include the smallest instance of each domain (which we use to select candidates and the configuration of \llmfixed{}) in the results of Table~\ref{tab:coverage}.
Thus, while we do not hold out the selection instance, this favours no column, because every system solves all seven of these instances, and they add one to every cell of their rows alike.

\begin{table}[t]
  \caption{Coverage (problems solved) per domain and in aggregate; every column is the virtual best over GBFS, A* and WA* except \llmfixed{}, which fixes one model and one search algorithm per domain in advance, and bold marks the best coverage in each row.}
  \label{tab:coverage}
  \centering
  \setlength{\tabcolsep}{3pt}
  \begin{tabular}{lrrrrrrrrrr}
    \toprule
    & & \multicolumn{4}{c}{\panda{}} & \multicolumn{5}{c}{\pytrich{}} \\
    \cmidrule(lr){3-6}\cmidrule(lr){7-11}
    Domain & Probs. & \rcff{} & \rclmcut{} & \rcadd{} & \pandalmcount{} & \tdg{} & \lmcount{} & \coiv{} & \llmfixed{} & \llmvb{} \\
    \midrule
    Barman                 & 20   & 18             & 15             & 16             & 18             & 18             & 19             & \textbf{20}                  & 18             & \textbf{20} \\
    Blocksworld            & 30   & 29             & 26             & \textbf{30}    & \textbf{30}    & 20             & \textbf{30}    & 28                           & 28             & 29 \\
    Depots                 & 30   & 27             & 23             & 22             & 29             & 23             & 27             & \textbf{30}                  & 26             & \textbf{30} \\
    Robot                  & 20   & \textbf{20}    & \textbf{20}    & \textbf{20}    & \textbf{20}    & 11             & 11             & 11                           & 11             & 11 \\
    Rover                  & 30   & 26             & 19             & 27             & 20             & \textbf{30}    & 16             & \textbf{30}                  & \textbf{30}    & \textbf{30} \\
    Towers                 & 20   & 14             & 13             & 14             & 14             & \textbf{15}    & \textbf{15}    & \textbf{15}                  & \textbf{15}    & \textbf{15} \\
    Lamps                  & 30   & 19             & 18             & 19             & 15             & 15             & 12             & 14                           & 15             & \textbf{20} \\
    \midrule
    Total                  & 180  & 153            & 134            & 148            & 146            & 132            & 130            & 148                          & 143            & \textbf{155} \\
    \bottomrule
  \end{tabular}
\end{table}

\llmfixed{} solves 143 of the 180 problems, thirteen more than \lmcount{} and eleven more than \tdg{}, and \llmvb{} raises the count to 155.
Both \pytrich{} baselines lose their coverage on the largest instances of each domain.
Outside Robot, the LLM heuristics solve all of these instances but one.
In aggregate \llmvb{} and \panda{}~\rcff{} are at parity, one problem apart, and they differ in where they solve.
\rcff{} solves Robot completely, while every \pytrich{} configuration, baseline or LLM, stops at the eleven smallest instances within the time limit.
On the other six domains \llmvb{} matches or exceeds \rcff{}, and \llmfixed{} matches or exceeds it on Barman, Rover and Towers.
The three largest Rover instances account for the aggregate margin, since no \panda{} configuration completes its preprocessing on them within the time limit, whereas \tdg{} and the LLM heuristics solve all three.
Lamps, which the prompts never saw and for which we wrote no hints, follows the same pattern, with the LLM heuristics at parity with the \panda{} engines and ahead of both \pytrich{} baselines, and the gain concentrated in a single model, Gemini~3~Flash.
The problems that separate \llmfixed{} from \llmvb{} are the price of committing to a model on a single small instance, which filters broken candidates but cannot rank the sound ones; we return to this limitation in Section~\ref{sec:discussion}.

\subsection{Search Efficiency and Plan Quality}
\label{sec:nodes}
\label{sec:planquality}

We compare search effort within \pytrich{}, where the planner, the grounding and the search code are the same and only the heuristic changes, pairing each instance that both sides solve and taking the virtual best over the three algorithms on both sides.
Against \tdg{}, \llmvb{} expands fewer nodes on 107 of the 132 shared instances (81\%), with a median node ratio of 0.74, and never expands more except on one Lamps instance.
Against \lmcount{} it wins 126 of 129 (98\%) at a median ratio of 0.46.
Figure~\ref{fig:scatter_expanded_nodes_tdg} shows the comparison against \tdg{}, where nearly every point lies below the diagonal.
The margin over \tdg{} depends on the domain.
It is largest in Barman, where \llmvb{} wins 17 of 18 instances at a median ratio of 0.02, followed by Lamps (13 of 15 at 0.24), Depots (21 of 23 at 0.41) and Blocksworld (20 of 20 at 0.45).
In Robot (10 of 11 at 0.77) and Towers (14 of 15 at 0.96) the margin shrinks, and in Rover the two sides tie on 18 of the 30 instances and \llmvb{} wins the other 12 by a handful of nodes.
The ordering follows the regularity of the decomposition hierarchy.
In Towers and Rover the task decomposition graph already yields a tight estimate and leaves little room for semantic guidance, whereas in Barman the ordering constraints of the methods hide most of the remaining work from \tdg{}.
Against \lmcount{} the advantage is uniform, with every domain won outright except Blocksworld (27 of 29).
Table~\ref{tab:domain-characteristics} in Appendix~\ref{app:pytrich-tdg} gives the structure of each domain beside its margin.
The node advantage belongs to the portfolio rather than to any configuration fixed in advance.
\llmfixed{} is at parity with \tdg{} (54 of 132 wins, median ratio 1.00) and ahead of \lmcount{} (78 of 126 at 0.93).
Figure~\ref{fig:main-pytrich-cactus} shows where its extra coverage lies.
Within ten thousand expanded nodes \llmfixed{} solves 91 problems, against 83 for \tdg{}, 86 for \lmcount{} and 116 for \llmvb{}.
Its curve stays close to those of the baselines on the smaller instances and separates on the larger ones, which it solves and they do not.

Node counts across planners are not directly comparable, since \panda{} and \pytrich{} ground differently and count different search nodes, so the node comparison against \panda{}~\rcff{} appears in Appendix~\ref{app:panda-rc2ff}, for completeness only.

Plan length, counted as primitive actions on both sides, separates the systems less than search effort does.
On the 143 instances shared with \panda{}~\rcff{}, \llmvb{} plans are shorter on 43, equal on 97 and longer on 3, with a median ratio of 1.00, and within \pytrich{} the plans of \llmvb{}, \tdg{} and \lmcount{} coincide on every shared instance outside Barman and Rover, where \llmvb{} is shorter (Appendix~\ref{app:planlength}).

\subsection{LLM Model Comparison}
\label{sec:models_results}

The main comparison covers five models: Claude Opus~4.6, Gemini 2.0 Flash, Gemini 2.5 Pro, Gemini 3 Flash and Gemini 3 Pro.
The remaining four models (GPT-4o, Claude Sonnet~4.5, GPT-5 and GPT-5.2) retained an executable heuristic in one domain each, so we exclude them from the per-domain tables.
No single model dominates.
Claude Opus~4.6 and Gemini 3 Flash each solve 148 of the 180 instances, Gemini 2.0 Flash 112, Gemini 3 Pro 100 and Gemini 2.5 Pro 82, and the virtual best is a portfolio, in which Claude Opus~4.6 carries the Blocksworld and Depots instances, Gemini 3 Pro most of Barman, and Gemini 3 Flash Rover, Towers and Lamps.
Over every solved run, the median expanded nodes range from 4{,}686 for Gemini 3 Flash and 6{,}599 for Claude Opus~4.6 to 14{,}908 for Gemini 3 Pro, 24{,}387 for Gemini 2.5 Pro and 45{,}294 for Gemini 2.0 Flash, a tenfold spread, wider than the effect of the search algorithm in Section~\ref{sec:algo_results}.
The generate-evaluate-select procedure compensates in part for weak models, since Gemini 2.0 Flash and Gemini 2.5 Pro contribute full coverage on the domains where their selected heuristics run without error.
On Robot, no LLM heuristic solves the nine largest instances within that limit, and neither does any \pytrich{} baseline.

Across the nine models and seven domains, we generated 928 candidate heuristics, between four and forty per model and domain.
Of these, 563 passed the import check that admits a candidate to the evaluation, and 482 solved at least one benchmark instance (Appendix~\ref{app:expdetails}).
Of the 365 candidates the import check rejected, 286 declare a constructor with a required argument, 201 of them a \texttt{model} argument that the interface passes to \texttt{initialize} instead, 73 fail to import, and 6 give the evaluation method the wrong signature.
The 81 candidates that imported but solved no instance fail on every run, most often by treating the bitwise state as a set, although the prompt specifies the representation; Appendix~\ref{app:exit-reasons} gives the exceptions and the exit reasons of the baselines' lost instances.

Four of the nine models have results in a single domain.
GPT-4o and Claude Sonnet~4.5 were run on Depots only, while GPT-5 and GPT-5.2 were run on all six IPC 2020 domains and passed the import check only on Towers.
We did not modify the prompt template to retry these failures, since doing so after observing test results would select against the very benchmark instances used for evaluation.
We therefore report a negative finding.
For GPT-5 and GPT-5.2, domain-specific prompting at the default sampling temperature produced no executable heuristic on five of the six domains.

\subsection{Effect of Search Algorithm}
\label{sec:algo_results}

Across all LLM runs, GBFS achieves the lowest median expanded nodes (8{,}761), followed by WA* (11{,}934) and A* (21{,}602), so GBFS is about $2.5\times$ more efficient than A* and $1.4\times$ more efficient than WA* (Figure~\ref{fig:algorithm_comparison}).
Paired per instance, GBFS expands fewer nodes than A* for every model, on 81 of 90 shared instances for Claude Opus~4.6, 110 of 134 for Gemini 3 Flash and 75 of 75 for Gemini 2.0 Flash.
The algorithm effect is largest for the strongest heuristics.
For Claude Opus~4.6 the median under A* is six times that under GBFS, against twice for Gemini 3 Flash and no gain for Gemini 2.5 Pro.

\section{Discussion}
\label{sec:discussion}

The pattern of results differs from the classical-planning case of \citet{CorreaPereiraSeipp2025}, where the LLM virtual best matches or exceeds state-of-the-art baselines across the board.
Here, the LLM virtual best reaches parity with \panda{}~\rcff{} in coverage, and its advantage shows within \pytrich{}, where it expands fewer nodes than \tdg{} on 81\% of shared instances and fewer than \lmcount{} on 98\%.
The two systems differ in the domains that they solve best. 
\rcff{} solves every Robot instance, whereas every \pytrich{} configuration, baseline or LLM, stops at the eleven smallest within the time limit, so the Robot gap seems to be a limit of the planner rather than of the heuristics.
On the other six domains the LLM virtual best matches or exceeds \rcff{}.
Plan length separates the systems less than search effort does.
In Towers every run of every system returns a plan of the same length, and in Blocksworld, Depots and Robot the best heuristic of each system returns plans of the same length on almost every shared instance, although weaker candidates return longer ones.
Only in Barman and Rover do the best heuristics disagree, and there the LLM heuristics return the shortest plans, ahead of both \tdg{} and \rcff{}.

The \hddl{} domain encoding provided in the prompt gives the LLM explicit access to the decomposition structure of the domain: method names, task hierarchies, and operator preconditions and effects.
This semantic richness allows the LLM to reason about which compound tasks are costly and which are cheap, producing estimates that are more informative than the purely syntactic relaxation that \tdg{} computes.
The gains over \tdg{} are largest where the cost of a task depends on the state, since \tdg{} estimates each pending task by its cheapest method and, in Barman, Lamps, Blocksworld and Depots, that method is the one that applies when the task is already achieved.
In Rover the methods fix the cost of each goal and the LLM virtual best only matches \tdg{}, and in Towers at most one method applies in any state, so no heuristic can matter.
Appendix~\ref{app:pytrich-tdg} goes through the methods of each domain, with Robot as the exception where neither heuristic guides the search.
Appendix~\ref{app:heuristics} lists a strong and a failed heuristic for Depots.

The tenfold spread in median expanded nodes across the five main models is the most striking model-level finding.
Newer models are not uniformly better, since Gemini 2.5 Pro solves fewer instances than Gemini 2.0 Flash and expands four times the nodes of Claude Opus~4.6 and Gemini 3 Flash.
The generate-evaluate-select procedure partially compensates for weak models, and even Gemini 2.0 Flash and Gemini 2.5 Pro contribute to full coverage on the domains where their selected heuristics run without error.
The efficiency gain from switching to GBFS grows with heuristic quality, as the median for Claude Opus~4.6 under A* is six times that under GBFS, while for Gemini 3 Flash it is twice.
This interaction suggests that discarding $g$-values is most beneficial when the heuristic itself encodes decomposition cost, since in that regime tracking path cost adds noise rather than signal.

\section{Related Work}
\label{sec:related}

\htn{} planning encodes domain knowledge in the decomposition hierarchy itself, complementing rather than replacing search heuristics~\citep{ErolHendlerNau1994,GhallabNauTraverso2004a}.
A line of work develops increasingly informative \htn{} heuristics, from the \tdg{} heuristic~\citep{BercherBehnkeHoellerEtAl2017} and the relaxed composition (RC) heuristics~\citep{HoellerBercherBehnkeEtAl2020} to landmark counting~\citep{HoellerBercher2021,PutrichMeneguzziPereira2025}; the progression-based \panda{} planner~\citep{HoellerBercherBehnkeEtAl2020} is the strongest system built on them.

LLMs perform poorly as planners~\citep{ValmeekamMarquezSreedharanEtAl2023}, and reliable use requires coupling them with sound search~\citep{TuisovVernikShleyfman2025,CorreaPereiraSeipp2025}.
In the \htn{} setting, \citet{XuMunozAvila2025} use LLMs to propose decompositions online, \citet{MunozAvilaAhaRizzo2025} interleave LLM-generated and symbolic decomposition, and \citet{OswaldSrinivasKokelEtAl2024} use them to author PDDL domains.
We use LLMs only to write the heuristic and delegate correctness to the search algorithm.

\citet{TuisovVernikShleyfman2025} argue that domain-independence may no longer be the right yardstick once LLMs can synthesise per-domain heuristics on demand.
Building on this, \citet{CorreaPereiraSeipp2025} prompt LLMs to write Python heuristics for classical planning, outperforming domain-independent baselines.
We adapt their pipeline to \htn{} planning, where the heuristic must reason over task networks and decomposition methods rather than flat states.

\section{Conclusion}
\label{sec:conclusion}

We investigate whether LLM-generated heuristics can serve as effective search guidance for \htn{} planning, extending the methodology of \citet{CorreaPereiraSeipp2025} from classical to hierarchical planning.
The answer is affirmative.
A configuration fixed in advance on one instance per domain solves 143 of 180 benchmark instances, against 132 for \tdg{} and 130 for \lmcount{} in the same planner, and the virtual best over all models and algorithms reaches parity with \panda{}~\rcff{}.
Within \pytrich{}, the LLM heuristics expand fewer nodes than \tdg{} on 81\% of the shared instances and \lmcount{} on 98\%.

The task hierarchy itself guides search, and \hypertension{} won the 2020 IPC \htn{} track with no explicit heuristic function~\citep{MagnaguagnoMeneguzziDeSilva2025}.
Prior work has focused on acquiring the methods themselves, from classical operators~\citep{Magnaguagno2017}, noisy observations~\citep{GrandFiorinoPellier2022}, or curriculum learning~\citep{LiNauRobertsEtAl2024}, reducing the manual effort of encoding the hierarchy.
We take the method library as given and generate a heuristic that exploits it.
\citet{TuisovVernikShleyfman2025} ask whether domain-independence remains a meaningful goal in the era of LLMs, and our results sharpen this question in the \htn{} setting, where the task hierarchy already encodes domain knowledge and the LLM heuristic adds a second layer of domain-specific guidance that exploits that encoding directly.

Our results come with four limitations.
The evaluation covers seven total-order \htn{} domains and leaves partial-order planning, numerical fluents and temporal domains open.
The virtual best chooses the model after seeing the result and is an upper bound; the configuration fixed in advance is the deployable one, and its single selection instance filters out broken candidates without ranking the sound ones.
Generation is stochastic, and every prompt includes the smallest and largest instance of its domain, so Appendix~\ref{app:prompt-context} recomputes every comparison without those instances, which confirms our results.
The LLM heuristics are Python programs written against the \pytrich{} interface and cannot run inside \panda{}, so we compare across planners primarily on coverage.

Extending the approach to partial-order \htn{} planning is the most important open direction, as the heuristic must then reason about the ordering space of the task network rather than a fixed sequence.
A domain-independent prompting condition, without the \hddl{} file, would determine how much of the observed gain derives from the domain knowledge supplied in the prompt and how much from the LLM's general reasoning about planning structure.
A prompt that asks for an estimate of the method applications still required, rather than of the missing primitive actions alone, would let the heuristic reason about the hierarchy itself.

\begin{ack}
We thank the anonymous reviewers for their comments, which led to the fixed configuration, the prompt-context recomputation and the cost analysis in this version.
After acceptance we repeated the full generation protocol for Claude Opus~4.6 with a larger API budget than was available at submission time, and every Claude Opus~4.6 result in this version comes from that batch.
We thank Anthropic for the Claude team plan for science, whose monthly API credit paid for that batch.
The experiments ran on the Maxwell high-performance computing cluster of the University of Aberdeen.
The authors declare no competing interests.
Felipe Meneguzzi acknowledges funding from Net Zero Technology Centre and the University of Aberdeen through their partnership in the UK National Decommissioning Centre.  
Victor Scherer Putrich acknowledges funding from Deutsche Forschungsgemeinschaft (DFG, German Research Foundation) - GRK 2853/1 ``Neuroexplicit Models of Language, Vision, and Action'' - project number 471607914.
André G.~Pereira acknowledges support from FAPERGS with projects 21/2551-0000741-9 and 25/2551-0002590-7.
This study was financed in part by the \textit{Coordenação de Aperfeiçoamento de Pessoal de Nível Superior -- Brasil} (CAPES) -- Finance Code~001.
\end{ack}

\bibliographystyle{abbrvnat}
\bibliography{references}

\clearpage
\appendix

\section{Prompt Templates}
\label{app:prompts}

\subsection{Base Prompt Structure}
\label{app:base-prompt}

The base prompt is a structured document of twelve numbered sections, preceded by a task preamble, delivered to the LLM for each domain.
Sections 1, 2 and 2.5 supply domain-specific information; sections 3--12 are fixed across all domains.

\begin{enumerate}
  \item \textbf{Preamble.}
    States the role (\emph{expert in hierarchical planning and heuristic design}), the target domain, and the required class name and parameter name for the generated Python class.
  \item \textbf{Section 1 --- Target domain definition.}
    The full \hddl{} domain file, presented verbatim in a fenced code block.
  \item \textbf{Section 2 --- Training set instances.}
    Two benchmark problems: the smallest (used for heuristic selection) and the largest available, both presented verbatim in \hddl{} format.
  \item \textbf{Section 2.5 --- Domain-specific hints.}
    An ablation condition rather than a pipeline input: the reported results use the prompt with this section empty, and Appendix~\ref{app:ablation} measures the effect of filling it with the block of the domain (Appendix~\ref{app:hints}).
\end{enumerate}

Sections 3--12 are identical across all domains and prompts.

\paragraph{Section 3 --- Grounded fact format.}
Explains that after grounding, facts are represented as \texttt{+predicate[arg1,arg2]} rather than in \hddl{} syntax, and provides a reference implementation of a parser:

\begin{lstlisting}[style=genpython,
  caption={Reference fact-name parser included verbatim in the prompt.}]
def _parse_fact_name(self, name: str):
    name = name.strip()
    if name.startswith("+") or name.startswith("-"):
        name = name[1:]
        bracket_idx = name.find("[")
        if bracket_idx > 0 and name.endswith("]"):
            predicate = name[:bracket_idx]
            args_str = name[bracket_idx+1:-1]
            args = args_str.split(",") if args_str else []
            return predicate, args
    return None, []
\end{lstlisting}

\paragraph{Section 4 --- Goal state access.}
Explains that \texttt{model.goals} is a bitwise integer, not a list of predicates, and shows the correct API call to retrieve goal fact names:

\begin{lstlisting}[style=genpython]
goal_fact_names = self.model.state_explicit_repr(self.model.goals)
# Returns: ['+communicated_soil_data[waypoint1]', ...]
\end{lstlisting}

\paragraph{Section 5 --- Bitwise state checks.}
Contrasts the correct O(1) method against the incorrect string-based alternative:

\begin{lstlisting}[style=genpython]
# CORRECT: O(1) bitwise check
is_true = (state_bitwise >> fact.global_id) & 1

# WRONG: O(n) string scan that causes timeouts on large instances
fact.name in self.model.state_explicit_repr(state)
\end{lstlisting}

\paragraph{Section 6 --- Required imports and method signatures.}
States the mandatory import paths and provides correct and incorrect variants of the two required method signatures:

\begin{lstlisting}[style=genpython]
from Pytrich.Heuristics.heuristic import Heuristic
from Pytrich.Search.htn_node import HTNNode
from Pytrich.model import Model

def initialize(self, model: Model, initial_node: HTNNode):
    h_value = self._compute(initial_node.state, initial_node.task_network)
    return super().initialize(model, h_value)

def __call__(self, parent_node: HTNNode, node: HTNNode) -> int:
    h_value = self._compute(node.state, node.task_network)
    super().update_info(h_value)
    return h_value
\end{lstlisting}

\paragraph{Section 7 --- Working example heuristic.}
A complete, working heuristic for a different domain, provided verbatim: the Towers heuristic for the other domains and a Snake heuristic, a domain outside the benchmark, for Towers.
It teaches the interface and not the domain, following \citet{CorreaPereiraSeipp2025}, and Appendix~\ref{app:ablation} shows that it improves code validity without changing search performance.
It shows how to parse grounded fact names and how to separate preprocessing in \texttt{initialize}, which builds lookup tables from the facts of the model, from per-node evaluation in \texttt{\_\_call\_\_}, which tests state bits and never iterates over the facts or operators of the model.

\paragraph{Sections 8 and 9 --- Class definitions.}
Full Python class definitions for \texttt{Fact}, \texttt{Operator}, \texttt{AbstractTask}, \texttt{Decomposition}, and \texttt{Model} are given verbatim so the LLM knows every attribute and method available at runtime.

\paragraph{Section 10 --- Admissibility vs.\ guidance.}
Tells the model to add its tie-breaking penalty to the maximum of the lower-bound terms, and that a penalty below one unit (e.g.\ $\times 0.001$) keeps the result a lower bound whenever the terms are, while still breaking symmetries.

\paragraph{Section 11 --- Winning and losing patterns.}
Six winning patterns (P1--P6) and four anti-patterns (A1--A4) are listed, each with a short code example.
The patterns record the failure modes we met while developing the prompt, and the runtime failures in Appendix~\ref{app:exit-reasons} are instances of anti-patterns A1 and A3.

\paragraph{Section 12 --- Required design procedure.}
The LLM is instructed to follow five steps and write the justification as top-of-file comments in the generated code: (1) name the domain bottleneck; (2) list two to three independent admissible lower bounds; (3) implement \texttt{initialize}; (4) implement \texttt{\_\_call\_\_} returning \texttt{max(term1, term2, \ldots)}; (5) add tie-breaking penalties if the domain has interchangeable parameters.
The LLM is asked to return a structured response (filename and code) to enable automated parsing.

\subsection{Iterative Refinement Prompt}
\label{app:refinement-prompt}

The refinement prompt appends a feedback section to the base prompt.
The section includes: the previous candidate code verbatim; a results table comparing the candidate against the \tdg{} baseline on the selection instance (expanded nodes, wall time, solution size, exit status); and auto-generated corrective guidance keyed to the observed failure mode.

Guidance is generated by inspecting the previous code for structural signatures:

\begin{itemize}
  \item \textbf{TIMEOUT}: If the code iterates \texttt{self.model.facts} or \texttt{self.model.operators} inside \texttt{\_\_call\_\_}, the feedback explicitly names this as anti-pattern A3 and instructs the LLM to move that work to \texttt{initialize}.
  \item \textbf{Worse than \tdg{}}: If the code never reads state bits (no \texttt{>{}>} or \texttt{\& 1} patterns), the feedback flags anti-pattern A1 and recommends adding state-aware per-task cost estimation.
    If no \texttt{max(} composition is present, the feedback recommends adding a second admissible lower bound.
  \item \textbf{Better than \tdg{}}: The feedback acknowledges the improvement and suggests tightening an existing term or adding a third \texttt{max} component.
  \item \textbf{Runtime error}: The error message is shown verbatim and the LLM is instructed to fix it.
\end{itemize}

The instruction \emph{``Do NOT start from scratch --- modify and improve the previous version''} is highlighted to prevent the LLM from discarding working structure when only a single component needs improvement.

\section{Details on Hints}
\label{app:hints}

The hint block is an ablation condition, not an input of the pipeline.
Every result in the main text comes from the prompt without it, and Appendix~\ref{app:ablation} measures its effect with Gemini~3 Flash, where it leaves coverage and code validity unchanged and reduces expanded nodes on about three quarters of the shared instances, mostly in Barman and Rover.
We wrote one block for each of the six IPC 2020 domains from the records of our development sessions, in which the model summarised the failure modes it had met and we kept the summary as hint text, so the blocks play the part of the hand-written checklist of \citet{CorreaPereiraSeipp2025}.
Lamps has no block.
Each block has three kinds of entries: representation caveats introduced by grounding, such as static facts that the grounder strips and method preconditions that it compiles into operators; the dominant search bottleneck of the domain; and guidance on lower bounds, symmetry-breaking terms and penalty scales.
The prompt introduces the block with the sentence ``These insights were discovered through extensive experimentation on this domain. Use them.''
The blocks follow verbatim.

\paragraph{Barman.}\mbox{}
\begin{lstlisting}[style=promptmd]
- BOTTLENECK: Parameter-binding symmetry. AchieveContainsShakerIngredient chooses which shot to use — all N shots are interchangeable, creating N-way branching per ingredient transfer (~N^(2K) for K cocktails)
- GROUNDING WARNING: In grounded operator names, the SHOT argument is at args[0], NOT args[1] as HDDL parameter order suggests. Always verify by examining grounded fact/operator names
- Goal-shot penalty: Penalize using goal shots (shots that appear in goal facts) as intermediaries. Penalty must be VERY large (~500/task) — small penalties (15-30) have zero effect because they are washed out by total h
- Working-shot ordering: Assign increasing penalties (0, 100, 200...) to break symmetry among identical working shots. STEP=100 >> STEP=50
- Hand preference: Penalize non-preferred hand (50/task) to halve branching
- Combined symmetry-breaking reduced pfile10 from 2.3M nodes (timeout) to 25K nodes (0.2s)
\end{lstlisting}

\paragraph{Blocksworld.}\mbox{}
\begin{lstlisting}[style=promptmd]
- BOTTLENECK: Stack clearing — to move a buried block, all blocks above must be unstacked first
- Key cost formulas: do_put_on(x,y) non-trivial = clear_cost(x) + clear_cost(y) + 3; clear_cost with stack height h = 1 + 2*h
- CRITICAL: Goal-stability / false-trivial detection. on(x,y) currently true but y is "unstable" (will be moved for another goal) means this goal is NOT free — it costs 5+. This single insight reduced p10 from 2.37M nodes to 5,813
- Nop actions and real (primitive) actions are completely disjoint in this domain, so you can SAFELY SUM admissible bounds on nop count and real action count (they never overlap)
- Multi-component max of 3 terms (nop+displacement, per-task recursive cost using stack heights, first-task exact cost) works best
\end{lstlisting}

\paragraph{Depots.}\mbox{}
\begin{lstlisting}[style=promptmd]
- GROUNDING WARNING: PANDA's grounder strips static facts like (place pallet0 depot0). To recover location info, parse grounded operator names: Drop[hoist0,crate1,pallet0,depot0] reveals pallet0 is at depot0
- GROUNDING WARNING: Method preconditions are compiled into __method_precondition_* operators, NOT stored in decomposition.pos_precons. Check for operators whose name starts with "__method_precondition" when analyzing decompositions
- Nop detection is non-trivial: "effectively nop" decompositions (all operators are effect-free) != single nop operator. Must check for effect-free operator chains
- Balanced additive combination max(goal_h, tn_h) + min(goal_h, tn_h)//2 works but can lose coverage on large instances; max() alone is safer
- Nop bonus: when a task's nop preconditions are NOT met (meaning the expensive path is required), add +2 to its cost
\end{lstlisting}

\paragraph{Robot.}\mbox{}
\begin{lstlisting}[style=promptmd]
- BOTTLENECK: Single-carry constraint — robot holds exactly one package at a time. Each delivery is a round-trip (go to package, pickup, go to goal, putdown)
- Inter-delivery travel bound: with P undelivered packages, the robot must make at least P pickup+putdown pairs (2P actions) PLUS navigate between them. Sum of min outgoing edges in the delivery graph minus max edge gives a Hamiltonian path lower bound
- Doors are one-way-open (once opened, stays open forever). Pre-compute all-pairs shortest paths ignoring door state, then add +1 per closed door on the direct connection as a correction
- Star topology: room "c" (corridor) connects to many rooms — pre-compute BFS distances from corridor hub
- For large instances (50+ packages), keep __call__ to O(P) not O(P^2) — Python overhead matters at 40K+ nodes/second
\end{lstlisting}

\paragraph{Rover.}\mbox{}
\begin{lstlisting}[style=promptmd]
- GROUNDING WARNING: PANDA's grounder STRIPS static facts (can_traverse, visible, at_lander, store_of, equipped_for_*). You CANNOT read rover capabilities from the state. Instead, extract static info from grounded operator names: navigate[rover0,wp1,wp2] tells you rover0 can traverse wp1→wp2
- TDG's weakness: it assumes minimum decomposition cost (rover already at target = nop). Your heuristic should compute ACTUAL navigation distance when rover is NOT at target: cost = 3*d + 1 per hop (visit + navigate + unvisit + action)
- For 2+ packages: use Hamiltonian path lower bound on inter-delivery travel (sum of min outgoing edges - max edge)
- Pre-compute rover navigation graph and all-pairs distances in initialize(). Do NOT pre-select "best" waypoints — that depends on runtime rover position
- Keep __call__ to O(goals) not O(goals * rovers). O(1) single-rover check per goal beats O(goals*rovers) iteration (14-24x slower, causes timeout)
\end{lstlisting}

\paragraph{Towers.}\mbox{}
\begin{lstlisting}[style=promptmd]
- Exact optimal cost for 3-peg Hanoi from any state is computable: 2^k - 1 for largest displaced ring k
- Process rings largest-to-smallest to avoid O(n^2) tower-tracing; build a peg-assignment array in O(n) during initialize()
- Memoize by state (bitwise int) — many decomposition paths reach the same state
- The heuristic example provided for this domain already demonstrates the optimal approach
\end{lstlisting}

\section{Prompt Ablation}
\label{app:ablation}

This appendix reports a prompt ablation.
It asks whether the search performance of the generated heuristics comes from the model or from the two domain-specific inputs the prompt supplies: the per-domain hint block (Section~2.5 of the prompt) and the worked example heuristic (Section~7).
All three arms use Gemini~3 Flash, twenty candidates per domain, one-shot, with provider thinking set to medium and temperature~1.0, on the six evaluation domains.
Every candidate that runs is evaluated under GBFS, A*, and WA* on all 150 instances of the six domains.
The arms differ in exactly one prompt input each relative to the first:
\begin{itemize}
  \item \textbf{Arm 1 (no example).} No domain hints: Section~2.5 is empty. No domain-specific example: Section~7 is replaced by a domain-agnostic stub that returns zero for every node, a syntactic template for the required signatures with no reasoning to imitate.
  \item \textbf{Arm 2 (hints).} Arm 1 plus the full hint block of the domain.
  \item \textbf{Arm 3 (fixed cross-domain example).} Arm 1 plus a working heuristic from a different domain as the worked example: the Towers heuristic for the other five domains, and a Snake heuristic for Towers.
\end{itemize}
The domain HDDL (Section~1 of the prompt) is present in all three arms and is not ablated.
Arms 2 and 3 are each compared with arm 1 only, since they differ from each other in two inputs at once.
The comparison is paired per instance: on the instances both arms solve, we count the share where the best candidate of the arm expands fewer nodes than the best candidate of arm 1, and the median of the per-instance node ratio.
Pairing controls for instance difficulty and is less sensitive to unequal candidate pools than a median over runs.
The result file is \href{https://github.com/meneguzzi-lab/Pytrich-LLM/blob/v1.0.4/Experiments/results/ablation.csv}{\texttt{Experiments/results/ablation.csv}} in the released repository~\citep{MeneguzziBuchweitzCorreaEtAl2026Software}, and the script \href{https://github.com/meneguzzi-lab/Pytrich-LLM/blob/v1.0.4/Experiments/python_scripts/generate_ablation_table.py}{\texttt{Experiments/python\_scripts/generate\_ablation\_table.py}} produces the tables from it.

\paragraph{Coverage.}
Table~\ref{tab:ablation-coverage} shows coverage per arm, as the virtual best over the three algorithms, beside the \llmvb{} column of Table~\ref{tab:coverage}.
Coverage saturates: most domains sit at the ceiling for every arm, so coverage cannot discriminate between them, and the signal lies in node counts.
The candidate pools are unequal because some generated heuristics fail to run; the pool sizes appear in parentheses.
The ablation candidates solve one Robot instance more than any candidate of the main study.

\begin{table}[h]
  \caption{Coverage per ablation arm, virtual best over GBFS, A*, and WA*, on the 150 instances of the six domains.
           Parentheses give the number of candidates, out of twenty, that run at all.
           \llmvb{} is the paper's virtual best over the five main models, for reference.}
  \label{tab:ablation-coverage}
  \centering
  \small
  \begin{tabular}{lrrrrr}
    \toprule
    Domain & Probs. & \llmvb{} & Arm 1 & Arm 2 & Arm 3 \\
    \midrule
    Barman-BDI        & 20  & 20  & 20 (15)  & 20 (20)  & 20 (20) \\
    Blocksworld-GTOHP & 30  & 29  & 29 (16)  & 30 (15)  & 30 (17) \\
    Depots            & 30  & 30  & 28 (18)  & 27 (16)  & 27 (16) \\
    Robot             & 20  & 11  & 12 (17)  & 12 (18)  & 12 (20) \\
    Rover-GTOHP       & 30  & 30  & 30 (16)  & 30 (11)  & 30 (14) \\
    Towers            & 20  & 15  & 15 (17)  & 15 (18)  & 15 (20) \\
    \midrule
    Total             & 150 & 135 & 134 (99) & 134 (98) & 134 (107) \\
    \bottomrule
  \end{tabular}
\end{table}

\paragraph{Search effort.}
Table~\ref{tab:ablation-nodes} reports the paired win rate and the median node ratio of arms 2 and 3 against arm 1 under each algorithm.
The two arms answer different questions.
The hints produce a search effect that does not depend on the algorithm, since arm 2 wins about three quarters of the shared instances under GBFS and WA* at a median node ratio near 0.8, and 59\% under A*.
The effect is concentrated in two domains, Barman-BDI and Rover-GTOHP, where the median ratio is about one third under GBFS, while Blocksworld-GTOHP, Depots, and Robot sit at parity.
The fixed cross-domain example leaves search performance unchanged, with arm 3 winning about half of the shared instances under GBFS and WA* at a median node ratio of 1.00 under both.
What the example buys is code validity, reported next.

\begin{table}[h]
  \caption{Paired comparison of arms 2 and 3 against arm 1 on expanded nodes, per algorithm.
           Win rate is the share of instances solved by both arms where the arm's best candidate expands fewer nodes than arm 1's; the denominator is the number of shared instances.
           Ratio is the median per-instance node ratio, arm over arm 1, so values below one favour the arm.}
  \label{tab:ablation-nodes}
  \centering
  \footnotesize
  \setlength{\tabcolsep}{3pt}
  \begin{tabular}{lrrrrrr}
    \toprule
    & \multicolumn{2}{c}{GBFS} & \multicolumn{2}{c}{A*} & \multicolumn{2}{c}{WA*} \\
    \cmidrule(lr){2-3}\cmidrule(lr){4-5}\cmidrule(lr){6-7}
    Domain & Arm 2 & Arm 3 & Arm 2 & Arm 3 & Arm 2 & Arm 3 \\
    \midrule
    \multicolumn{7}{l}{\emph{Win rate against arm 1}} \\
    Barman-BDI        & 20/20  & 13/20 & 10/10 & 10/10 & 19/19  & 19/19 \\
    Blocksworld-GTOHP & 26/29  & 7/29  & 12/15 & 11/16 & 22/25  & 10/26 \\
    Depots            & 12/27  & 25/27 & 13/20 & 14/20 & 13/27  & 25/27 \\
    Robot             & 1/12   & 0/12  & 5/12  & 5/12  & 1/12   & 2/12 \\
    Rover-GTOHP       & 30/30  & 22/30 & 5/9   & 3/9   & 30/30  & 3/29 \\
    Towers            & 13/15  & 0/15  & 3/15  & 0/15  & 13/15  & 0/15 \\
    Total             & 102/133 (77\%) & 67/133 (50\%) & 48/81 (59\%) & 43/82 (52\%) & 98/128 (77\%) & 59/128 (46\%) \\
    \midrule
    \multicolumn{7}{l}{\emph{Median node ratio against arm 1}} \\
    Barman-BDI        & 0.32 & 0.99 & 0.00 & 0.94 & 0.04 & 0.71 \\
    Blocksworld-GTOHP & 0.97 & 1.04 & 0.89 & 0.90 & 0.94 & 1.04 \\
    Depots            & 1.02 & 0.82 & 0.87 & 0.99 & 1.02 & 0.84 \\
    Robot             & 1.00 & 1.00 & 1.00 & 1.00 & 1.00 & 1.00 \\
    Rover-GTOHP       & 0.30 & 0.92 & 0.98 & 1.22 & 0.29 & 1.25 \\
    Towers            & 0.65 & 1.00 & 1.00 & 1.00 & 0.70 & 1.00 \\
    Total             & 0.78 & 1.00 & 0.96 & 0.99 & 0.81 & 1.00 \\
    \bottomrule
  \end{tabular}
\end{table}

\paragraph{Code validity.}
Table~\ref{tab:ablation-validity} separates candidates that fail to compile from candidates that compile and then fail at run time, measured by running each of the 360 candidates on one small instance of its domain.
The worked example cuts runtime failures by more than half, from 17.5\% and 19.2\% to 7.5\%, while compile failures rise from one in arm 1 and none in arm 2 to five in arm 3.
Of the 53 runtime failures across the three arms, 28 are attribute errors, all in arms 1 and 2, and 16 of these read the model before the base-class initialiser assigns it, which is the mistake the worked example shows how to avoid.
The example teaches the \pytrich{} API rather than better Python.
The hints do not help validity: arm 2 is marginally worse than arm 1 on both counts, so whatever the hints buy is search guidance.

\begin{table}[h]
  \caption{Code validity per arm over the 120 candidates each arm generates, six domains by twenty.
           A candidate is usable if it imports and completes a run on one small instance of its domain.}
  \label{tab:ablation-validity}
  \centering
  \small
  \begin{tabular}{lrrrrr}
    \toprule
    Arm & Generated & Usable & Fails to compile & Fails at run time & Failure rate \\
    \midrule
    Arm 1 (no example)            & 120 & 98  & 1 (0.8\%) & 21 (17.5\%) & 18.3\% \\
    Arm 2 (hints)                 & 120 & 97  & 0 (0.0\%) & 23 (19.2\%) & 19.2\% \\
    Arm 3 (fixed cross-dom.\ ex.) & 120 & 106 & 5 (4.2\%) & 9 (7.5\%)   & 11.7\% \\
    \bottomrule
  \end{tabular}
\end{table}

\paragraph{What the ablation cannot show.}
Three limits apply.
First, A* is incomplete for total-order \htn{} planning, so the A* columns rest on 82 shared instances rather than 133 and are not comparable to the GBFS figures.
Second, twenty candidates per domain can only show the absence of a detectable effect, and we do not measure the variance between independent draws of twenty candidates from the same arm.
This applies to arm 3's ratio of 1.00 as much as to the domains where arm 2 is at parity with arm 1.
Third, usable on one small instance is a weak bar: it establishes that the code executes, not that the heuristic is informative, and a candidate that crashes only on large instances counts as usable here.

\section{Experimental Details}
\label{app:expdetails}

\subsection{Models Used}

Table~\ref{tab:models} lists all evaluated models.

\begin{table}[h]
  \centering
  \caption{LLM models evaluated.
           \textsuperscript{$\dagger$}Run on Depots only.
           \textsuperscript{$\ddagger$}Passed the import check on Towers only.
           \textsuperscript{$\S$}No Depots candidate loaded (Table~\ref{tab:funnel}).}
    \label{tab:models}
  \begin{tabular}{lll}
    \toprule
    Key in results                            & Model name        & Provider               \\
    \midrule
    ClaudeOpus46                              & Claude Opus~4.6   & Anthropic \\
    ClaudeSonnet45\textsuperscript{$\dagger$} & Claude Sonnet~4.5 & Anthropic \\
    Gemini20Flash                             & Gemini 2.0 Flash  & Google    \\
    Gemini25Pro                               & Gemini 2.5 Pro    & Google    \\
    Gemini3Flash                              & Gemini 3 Flash    & Google    \\
    Gemini3Pro\textsuperscript{$\S$}          & Gemini 3 Pro      & Google    \\
    ChatGPT4o\textsuperscript{$\dagger$}      & GPT-4o            & OpenAI    \\
    GPT5\textsuperscript{$\ddagger$}          & GPT-5             & OpenAI    \\
    GPT52\textsuperscript{$\ddagger$}         & GPT-5.2           & OpenAI    \\
    \bottomrule
  \end{tabular}
\end{table}

\subsection{Generation settings}

For candidate generation, we use provider-default sampling settings and cap responses at 16,384 tokens.
Each model-domain pair generates its candidates by independent one-shot invocations of the prompt, evaluated and selected as described in Section~\ref{sec:selection}.
For Claude Opus~4.6 no thinking budget is sent and the temperature is the provider default, which the model does not let us set.
Table~\ref{tab:funnel} gives, for every model and domain, the number of candidates generated, the number that pass the import check of the task-list generator (a candidate is admitted if its file imports, defines a \texttt{Heuristic} subclass with the two required methods, and has a constructor without required arguments), and the number that solve at least one benchmark instance.

\begin{table}[htbp]
  \centering
  \caption{Candidate funnel per model and domain, as the number generated, the number that passed the import check, and the number that solved at least one instance.
           A dash marks a pair for which no candidates were generated; $\dagger$ marks the one importable Gemini 3 Pro Depots candidate, whose registry entry was disabled, so it never loaded.}
  \label{tab:funnel}
  \footnotesize
  \setlength{\tabcolsep}{3pt}
  \begin{tabular}{lrrrrrrrr}
\toprule
Model & Barman & Blocksworld & Depots & Robot & Rover & Towers & Lamps & Total \\
\midrule
Claude Opus~4.6 & 20/20/20 & 20/17/13 & 20/19/19 & 20/19/14 & 20/20/20 & 20/20/20 & 20/20/20 & 140/135/126 \\
Gemini 2.0 Flash & 29/29/27 & 39/34/12 & 40/37/35 & 39/39/37 & 40/39/15 & 39/37/26 & -- & 226/215/152 \\
Gemini 2.5 Pro & 20/11/11 & 20/14/13 & 10/7/6 & 20/13/13 & 6/6/5 & 29/20/19 & 20/20/20 & 125/91/87 \\
Gemini 3 Flash & 6/6/6 & 6/6/6 & 6/6/6 & 6/6/6 & 6/6/6 & 6/6/6 & 20/19/19 & 56/55/55 \\
Gemini 3 Pro & 25/5/5 & 22/2/2 & 20/1/0$^\dagger$ & 22/2/2 & 21/1/1 & 21/18/18 & -- & 131/29/28 \\
GPT-4o & -- & -- & 4/4/4 & -- & -- & -- & -- & 4/4/4 \\
Claude Sonnet~4.5 & -- & -- & 6/6/6 & -- & -- & -- & -- & 6/6/6 \\
GPT-5 & 20/0/0 & 20/0/0 & 20/0/0 & 20/0/0 & 20/0/0 & 20/20/20 & -- & 120/20/20 \\
GPT-5.2 & 20/0/0 & 20/0/0 & 20/0/0 & 20/0/0 & 20/0/0 & 20/8/4 & -- & 120/8/4 \\
\bottomrule
\end{tabular}

\end{table}

\subsection{Execution details}

All jobs are executed through SLURM job arrays with one benchmark instance per task.
Each task is allocated 1 CPU core, 8~GB RAM, and a 30-minute wall-clock limit.
Both \pytrich{} and \panda{} runs use the same resource limits.
All \panda{} runs use one build of \texttt{pandaPIengine}, with seed 42 for its tie-breaker and path cost counted per action; the released repository builds the parser, grounder and engine from pinned commits.
Every run executes on one core of an Intel Xeon Gold 6138 processor in a shared SLURM partition.

\subsection{Exit reasons}
\label{app:exit-reasons}

We read the exit of every unsolved run from its log: the memory limit of 8~GB, the time limit of 30 minutes, or a Python exception.
Every unsolved run falls into one of the three.
Table~\ref{tab:exit-reasons} gives the exits of the two \pytrich{} baselines on the instances they fail, all three search algorithms counted.
\tdg{} fails 48 instances, 123 of its 144 runs on them at the memory limit and 21 at the time limit, and \lmcount{} fails 50, 140 of 150 runs at the memory limit and 10 at the time limit; neither raises an exception.
The exceptions are \tdg{} on Barman and most of Towers, and \lmcount{} on Rover p17, which time out.
So the baselines lose almost every instance they miss to memory, with all three runs at the limit on 39 of the 48 \tdg{} misses and 46 of the 50 \lmcount{} misses.
On the same 65 instances, the five main models' candidates exhaust memory on most of their unsolved runs as well, with time the most common exit for Claude Opus~4.6 alone; their coverage gain comes from the instances they solve, chiefly in Rover, Blocksworld and Depots, and no run of any model solves Robot pfile\_10\_020 and larger or Towers 16 to 20.

\begin{table}[htbp]
  \centering
  \caption{Exits of the \pytrich{} baselines on the instances they fail, over the three search algorithms, as runs that end at the memory limit (M), at the time limit (T), and the number of instances on which all three runs end at the memory limit (all M).}
  \label{tab:exit-reasons}
  \footnotesize
  \begin{tabular}{lrrrrrrrr}
\toprule
& \multicolumn{4}{c}{\tdg{}} & \multicolumn{4}{c}{\lmcount{}} \\
\cmidrule(lr){2-5} \cmidrule(lr){6-9}
Domain & Unsolved & M & T & all M & Unsolved & M & T & all M \\
\midrule
Barman & 2 & 0 & 6 & 0 & 1 & 3 & 0 & 1 \\
Blocksworld & 10 & 28 & 2 & 9 & 0 & 0 & 0 & 0 \\
Depots & 7 & 20 & 1 & 6 & 3 & 9 & 0 & 3 \\
Robot & 9 & 27 & 0 & 9 & 9 & 27 & 0 & 9 \\
Rover & 0 & 0 & 0 & 0 & 14 & 39 & 3 & 13 \\
Towers & 5 & 3 & 12 & 0 & 5 & 8 & 7 & 2 \\
Lamps & 15 & 45 & 0 & 15 & 18 & 54 & 0 & 18 \\
\midrule
Total & 48 & 123 & 21 & 39 & 50 & 140 & 10 & 46 \\
\bottomrule
\end{tabular}

\end{table}

Of the 563 candidates that pass the import check, 81 solve no instance.
Five of them never loaded: four GPT-5.2 Towers candidates and one Gemini 3 Pro Depots candidate have their registry entries disabled, so every run fails on the lookup before the file is read.
The other 76 raise the same exception on every one of their runs, and none times out.
In 43 the candidate treats the state or the goal, a bitwise integer, as a set or a list, most often by iterating over \texttt{model.goals}.
In 13 it reads an attribute the search node does not have, such as \texttt{uncompleted\_tasks} or \texttt{method}.
Six fail on an import path or an undefined name, and 14 on the candidate's own logic, such as splitting a fact name on a space it does not contain.
By model, Gemini 2.0 Flash accounts for 63 of the 76, Claude Opus~4.6 for 9 and Gemini 2.5 Pro for 4.
No run in the record used a file that fails today's import check, so the two failure stages do not overlap.

\subsection{Prompt-context instances}
\label{app:prompt-context}

Every prompt contains the smallest and the largest instance of its domain, and both are in the evaluated set.
Table~\ref{tab:prompt-context} recomputes coverage with those twelve instances removed, 168 remaining.
No comparison changes: \llmvb{} and \rcff{} stay at parity, \llmfixed{} stays ahead of both \pytrich{} baselines, and the ordering of every column is the same as in Table~\ref{tab:coverage}.

\begin{table}[htbp]
  \centering
  \caption{Coverage with the twelve prompt-context instances removed (168 instances); columns as in Table~\ref{tab:coverage}.}
  \label{tab:prompt-context}
  \small
  \begin{tabular}{lrrrrrrrr}
\toprule
Domain & Probs. & \rcff{} & \rclmcut{} & \pandalmcount{} & \tdg{} & \lmcount{} & \llmfixed{} & \llmvb{} \\
\midrule
Barman & 18 & 17 & 14 & 17 & 17 & 18 & 17 & 18 \\
Blocksworld & 28 & 28 & 25 & 28 & 19 & 28 & 27 & 27 \\
Depots & 28 & 26 & 22 & 28 & 22 & 26 & 25 & 28 \\
Robot & 18 & 18 & 18 & 18 & 10 & 10 & 10 & 10 \\
Rover & 28 & 25 & 18 & 19 & 28 & 15 & 28 & 28 \\
Towers & 19 & 13 & 12 & 13 & 14 & 14 & 14 & 14 \\
Lamps & 29 & 18 & 17 & 14 & 14 & 11 & 14 & 19 \\
\midrule
Total & 168 & 145 & 126 & 137 & 124 & 122 & 135 & 144 \\
\bottomrule
\end{tabular}

\end{table}

\subsection{Heuristic cost}
\label{app:cost}

Table~\ref{tab:cost} separates the once-per-problem preprocessing cost of each heuristic from its per-node evaluation cost, measured on the GBFS runs of \tdg{}, \lmcount{} and the selected candidate of each model, all on one core of the same processor; Lamps was not timed.
The prompt asks that each term of the per-node evaluation consist of constant-time bit tests and that every loop over the facts or operators of the model move to preprocessing, and the generated heuristics do keep preprocessing cheaper than the baselines in every domain, since the baselines build decomposition graphs and landmark structures while the generated code builds small lookup tables.
The generated code does not always comply with this request.
Per-node evaluation is comparable to the baselines, at tens of microseconds, but Table~\ref{tab:cost-scaling} shows that the cost of the selected Claude Opus~4.6 heuristic on Blocksworld grows with the instance, from 37 to 261 microseconds per node, while \lmcount{} stays near 20.

\begin{table}[htbp]
  \centering
  \caption{Median preprocessing cost (init, milliseconds) and per-node evaluation cost (microseconds) per domain under GBFS, LLM heuristics pooled over the five main models.}
  \label{tab:cost}
  \small
  \begin{tabular}{lrrrrrr}
\toprule
 & \multicolumn{2}{c}{LLM heuristics} & \multicolumn{2}{c}{\tdg{}} & \multicolumn{2}{c}{\lmcount{}} \\
\cmidrule(lr){2-3}\cmidrule(lr){4-5}\cmidrule(lr){6-7}
Domain & init (ms) & per node ($\mu$s) & init (ms) & per node ($\mu$s) & init (ms) & per node ($\mu$s) \\
\midrule
Barman & 8.2 & 28 & 93 & 17 & 157 & 15 \\
Blocksworld & 2.7 & 37 & 27 & 19 & 19 & 15 \\
Depots & 1.6 & 26 & 43 & 17 & 74 & 13 \\
Robot & 0.1 & 11 & 1.1 & 11 & 1.0 & 11 \\
Rover & 4.7 & 49 & 2{,}607 & 53 & 6{,}254 & 19 \\
Towers & 3.6 & 13 & 253 & 18 & 1{,}071 & 12 \\
\bottomrule
\end{tabular}

\end{table}

\begin{table}[htbp]
  \centering
  \caption{Per-node evaluation cost on Blocksworld instances of increasing size under GBFS.}
  \label{tab:cost-scaling}
  \small
  \begin{tabular}{lrrrrrr}
\toprule
Instance & p01 & p05 & p10 & p14 & p17 & p19 \\
\midrule
\lmcount{} per node ($\mu$s) & 19 & 12 & 17 & 19 & 20 & 21 \\
Claude Opus~4.6 per node ($\mu$s) & 37 & 43 & 196 & 169 & 254 & 261 \\
\bottomrule
\end{tabular}

\end{table}

\subsection{Model-specific failures}

GPT-4o and Claude Sonnet~4.5 were run on Depots only.
GPT-5 and GPT-5.2 were run on the six IPC 2020 domains and passed the import check only on Towers; on the other five domains every candidate declares a constructor that takes the model as a required argument, which the interface passes to \texttt{initialize} instead, or fails to import.
Gemini 3 Pro passed the import check with one to five candidates on four domains and eighteen on Towers; its one importable Depots candidate never loaded (Table~\ref{tab:funnel}).
These outcomes are included in the aggregate accounting and discussed in Section~\ref{sec:models_results}.

\section{Generated Heuristic Examples}
\label{app:heuristics}

This appendix lists two heuristics from the dataset of record for Depots, the domain in which Claude Opus~4.6 carries the virtual best.
The first is the candidate excerpted in Figure~\ref{fig:prompt-excerpt}, selected from the twenty Claude Opus~4.6 candidates on the smallest instance; it solves all 30 Depots instances and expands the fewest nodes of any LLM run on 10 of them.
The second is a Gemini 2.5 Pro candidate and shows the failure mode that dominates the weaker models on this domain.

\subsection{Strong Heuristic: Claude Opus~4.6 on Depots}
\label{app:good-heuristic}

The heuristic prices every pending task from the current state.
A pending \texttt{do\_put\_on} task costs nothing when its crate is already on the target surface, which is the case in which \tdg{} also returns the cost of the empty \texttt{nop} method, and otherwise costs the actions needed to clear the target, clear the crate, and move it, with the move priced by whether the crate is at the same place, at another place, in a truck or on a hoist.
The final value is the maximum of three terms: the summed cost of the unsatisfied goals, the summed cost of the pending tasks, and the number of unsatisfied goals plus a clearing term for goal surfaces that the task network does not yet account for.
All fact-id maps are built once in \texttt{initialize}, so each evaluation only tests bits.
The listing abridges \texttt{initialize} and omits the helper predicates (\texttt{\_is\_clear}, \texttt{\_count\_stack\_above}, \texttt{\_get\_crate\_location}, \texttt{\_is\_crate\_in\_truck}, \texttt{\_is\_crate\_being\_lifted}, \texttt{\_get\_surface\_place}, \texttt{\_is\_on\_goal}), each of which is one or two bit tests over the maps built in \texttt{initialize}.

\begin{lstlisting}[style=genpython,
  caption={Selected heuristic for Depots (Claude Opus~4.6, candidate 9), abridged.},
  label={lst:depots-good}]
"""Domain-Dependent Heuristic for Depots HDDL Domain

Bottleneck Analysis:
    The primary bottleneck in Depots is the need to clear crates stacked on top
    of target surfaces before placing goal crates, combined with the need to
    transport crates between different places using trucks (which requires
    lift/load/drive/unload/drop sequences). [...]

Admissible Lower Bound Terms:
    1. Goal-displacement cost: For each unsatisfied goal (on ?c ?s), we need at
       minimum 1 action (Drop) if the crate is already lifted at the right place,
       2 actions (Lift+Drop) if at the same place on wrong surface, or ~5 actions
       (Lift+Load+Drive+Unload+Drop) if at a different place. [...]
    2. Clearing cost: For each goal surface that needs to be cleared, we need
       at least 2 actions per blocking crate (Lift+Load minimum). [...]
    3. Task network lower bound: Count remaining compound tasks weighted by
       minimum decomposition depth. [...]
"""

class DepotsAnthropicOpus46Heuristic9(Heuristic):

    def initialize(self, model: Model, initial_node: HTNNode):
        """Preprocess: build fact maps, goal structures, place assignments."""
        self.model = model
        self.on_fact_id = {}      # (crate, surface) -> global_id
        self.at_fact_id = {}      # (locatable, place) -> global_id
        self.in_fact_id = {}      # (crate, truck) -> global_id
        self.clear_fact_id = {}   # surface -> global_id
        self.lifting_fact_id = {} # (hoist, crate) -> global_id
        for f in model.facts:
            pred, args = self._parse_fact_name(f.name)
            if pred == "on" and len(args) == 2:
                self.on_fact_id[(args[0], args[1])] = f.global_id
            [...]
        self.goal_on = {}  # crate -> target_surface
        for gname in model.state_explicit_repr(model.goals):
            pred, args = self._parse_fact_name(gname)
            if pred == "on" and len(args) == 2:
                self.goal_on[args[0]] = args[1]
        [...]
        self.task_info = {}  # abstract_task.global_id -> (task_name, args)
        for at in model.abstract_tasks:
            [...]
        h_value = self._compute(initial_node.state, initial_node.task_network)
        return super().initialize(model, h_value)

    def _estimate_put_on_cost(self, crate, target_surface, state):
        """Estimate minimum cost for do_put_on(crate, target_surface)."""
        key = (crate, target_surface)
        if key in self.on_fact_id and (state >> self.on_fact_id[key]) & 1:
            return 0  # Already on target - nop
        cost = 0
        if not self._is_clear(target_surface, state):
            blocking = self._count_stack_above(target_surface, state)
            cost += blocking * 4  # Lift + Load + Unload + Drop per crate (conservative)
        target_place = self._get_surface_place(target_surface)
        crate_place = self._get_crate_location(crate, state)
        if crate_place is not None:
            if not self._is_clear(crate, state):
                blocking_on_crate = self._count_stack_above(crate, state)
                cost += blocking_on_crate * 4
            if crate_place == target_place:
                cost += 2   # Same place: Lift + Drop
            else:
                cost += 5   # Different place: Lift + Load + Drive + Unload + Drop
        elif self._is_crate_in_truck(crate, state):
            cost += 3       # Drive + Unload + Drop
        elif self._is_crate_being_lifted(crate, state):
            cost += 1       # Drop
        else:
            cost += 5       # Unknown state - conservative estimate
        return cost

    def _compute(self, state, task_network):
        """Compute heuristic value for a state and task network."""
        if len(task_network) == 0:
            if self.model.goals & state == self.model.goals:
                return 0
            return float('inf')
        # Term 1: Goal displacement
        goal_cost = 0
        unsatisfied_goals = 0
        for crate, target in self.goal_on.items():
            if not self._is_on_goal(crate, state):
                unsatisfied_goals += 1
                [...]   # 2 at the same place, 5 elsewhere, 3 in a truck, 1 on a hoist
        # Term 2: Task network cost estimation
        task_cost = 0
        for task in task_network:
            gid = task.global_id
            if gid >= self.model.iabt_init and gid <= self.model.iabt_end:
                task_name, args = self.task_info[gid]
                if task_name == "do_put_on" and len(args) == 2:
                    task_cost += self._estimate_put_on_cost(args[0], args[1], state)
                elif task_name == "do_clear" and len(args) == 2:
                    if not self._is_clear(args[0], state):
                        blocking = self._count_stack_above(args[0], state)
                        task_cost += max(blocking * 4, 2)
                elif task_name == "do_get_truck" and len(args) == 2:
                    at_key = (args[0], args[1])
                    if at_key in self.at_fact_id:
                        if not ((state >> self.at_fact_id[at_key]) & 1):
                            task_cost += 1  # Drive
                elif task_name == "do_lift_crate":
                    task_cost += 1
                elif task_name in ("do_load_truck", "do_unload_truck"):
                    task_cost += 2
                else:
                    task_cost += 1
            elif gid >= self.model.iop_init and gid <= self.model.iop_end:
                op = self.model.operators[gid - self.model.iop_init]
                if "nop" not in op.name:
                    task_cost += 1
        # Term 3: Clearing cost for goal surfaces not yet in the task network
        clearing_cost = 0
        for crate, target in self.goal_on.items():
            if not self._is_on_goal(crate, state) and not self._is_clear(target, state):
                top_crate = self._get_crate_on_surface(target, state)
                if top_crate is not None and top_crate != crate:
                    if self.goal_on.get(top_crate) != target:
                        clearing_cost += 2  # At least Lift+Load to move it
        return max(goal_cost, task_cost, unsatisfied_goals + clearing_cost)

    def __call__(self, parent_node: HTNNode, node: HTNNode) -> int:
        h_value = self._compute(node.state, node.task_network)
        super().update_info(h_value)
        return h_value
\end{lstlisting}

\subsection{Weak Heuristic: Gemini 2.5 Pro on Depots}
\label{app:bad-heuristic}

The heuristic precomputes a cost for every method by a fixed point over the action names in its subtasks, which is a reasonable plan, and then evaluates a node by looking up the method that produced it.
It reads that method from an attribute \texttt{node.method} that the \pytrich{} search node does not define, so the guard on the first line of \texttt{\_calculate\_node\_heuristic} fires on every node and the heuristic returns zero throughout the search, which is then blind.
The record shows the consequence.
Forty-four Depots candidates, all 35 from Gemini 2.0 Flash, all 6 from Gemini 2.5 Pro and 3 of the 4 from ChatGPT-4o, expand exactly the same number of nodes as one another on every instance, and solve the same 17.
Most of them read the same missing attribute; the rest never read the state at all.
This is the failure mode behind anti-pattern A1 of the prompt (Appendix~\ref{app:prompts}), and the selection step cannot catch it, since a blind search solves the selection instance.

\begin{lstlisting}[style=genpython,
  caption={Weak heuristic for Depots (Gemini 2.5 Pro, candidate 3), abridged.},
  label={lst:depots-bad}]
class DepotGemini25ProHeuristic3(Heuristic):
    def __init__(self, depot_gemini25pro_heuristic_3=None, drive_penalty=3,
                 manipulation_penalty=1, clear_penalty=0.5):
        [...]
        self.method_costs = {}
        self.task_costs = {}

    def initialize(self, model: Model, initial_node: HTNNode):
        self.model = model
        self._precompute_costs()
        h_value = 0
        return super().initialize(model, h_value)

    def _precompute_costs(self):
        """Pre-computes the heuristic cost for each decomposition method and
        abstract task using a fixed-point iteration approach."""
        for task in self.model.abstract_tasks:
            self.task_costs[task.name] = float('inf')
        changed = True
        while changed and iterations < max_iterations:
            changed = False
            for method in self.model.decompositions:
                current_method_cost = 0
                for subtask in method.task_network:
                    if isinstance(subtask, Operator):
                        action_name = subtask.name.lower().split(' ')[0]
                        if 'drive' in action_name:
                            current_method_cost += self.drive_penalty
                        elif any(op in action_name
                                 for op in ['lift', 'drop', 'load', 'unload']):
                            current_method_cost += self.manipulation_penalty
                    elif isinstance(subtask, AbstractTask):
                        [...]
                self.method_costs[method.name] = current_method_cost
            [...]   # task cost = min over its methods

    def _calculate_node_heuristic(self, node: HTNNode) -> float:
        """The value is based on the pre-computed cost of the method used for
        decomposition."""
        if not hasattr(node, 'method') or not node.method:
            return 0.0
        return float(self.method_costs.get(node.method.name, 1.0))

    def __call__(self, parent_node: HTNNode, node: HTNNode) -> float:
        h_value = self._calculate_node_heuristic(node)
        super().update_info(h_value)
        return h_value
\end{lstlisting}

The two listings differ on the dimension that Section~\ref{sec:discussion} identifies.
The first prices each pending task from the state and so sees when a task is already achieved or nearly so, while the second computes a static cost per method and then fails to attach it to the search node, so its estimate carries no information at all.

\section{Additional Results and Figures}
\label{app:fullresults}

This appendix contains the complete set of comparison figures for the four baselines.
The main paper includes the within-\pytrich{} scatter plots and the algorithm-comparison chart; the cross-planner figures against \panda{}~\rcff{} appear here.
All figures use the \llmvb{} virtual best over the five main models and three search algorithms.

\FloatBarrier
\subsection{\panda{}~\rcff{}}
\label{app:panda-rc2ff}

\begin{figure}[ht]
  \centering
  \includegraphics[width=0.48\linewidth]{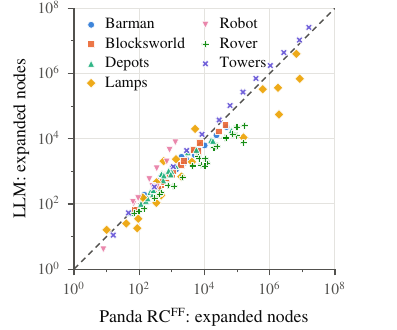}
  \caption{Cross-planner search effort: per-problem expanded nodes of \llmvb{} against \panda{}~\rcff{}; points below the diagonal indicate an \llmvb{} advantage.
           Node counts are planner-specific, since \panda{} and \pytrich{} differ in per-node bookkeeping, and should be read with care across systems.
           On the 143 shared instances \llmvb{} expands fewer nodes on 92 (64\%), median ratio 0.88, carried by Rover, Blocksworld and Lamps; Robot and Towers are losses.
           Table~\ref{tab:margins-rc2ff} gives the margins per domain.}
  \label{fig:scatter_expanded_nodes}
\end{figure}

\begin{table}[ht]
  \centering
  \caption{Per-domain margins on expanded nodes, \llmvb{} against \panda{}~\rcff{}, on the instances both solve: how many instances each side expands fewer nodes on, the median per-instance node ratio, and the mean reduction in nodes on the instances \llmvb{} wins.}
  \label{tab:margins-rc2ff}
  \small
  \begin{tabular}{lrrrrr}
\toprule
Domain & Shared & \llmvb{} fewer & \rcff{} fewer & Median ratio & Mean reduction on wins (\%) \\
\midrule
Barman & 18 & 10 & 8 & 0.94 & 27 \\
Blocksworld & 28 & 26 & 2 & 0.89 & 15 \\
Depots & 27 & 14 & 13 & 0.98 & 24 \\
Robot & 11 & 2 & 9 & 1.80 & 27 \\
Rover & 26 & 26 & 0 & 0.26 & 69 \\
Towers & 14 & 1 & 13 & 1.56 & 31 \\
Lamps & 19 & 13 & 6 & 0.47 & 67 \\
\midrule
All & 143 & 92 & 51 & 0.88 & 41 \\
\bottomrule
\end{tabular}

\end{table}

\begin{figure}[ht]
  \centering
  \begin{subfigure}[t]{0.48\textwidth}
    \centering
    \includegraphics[width=\linewidth]{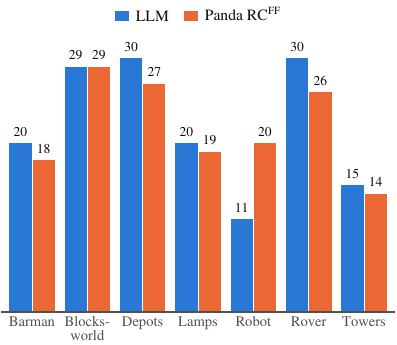}
    \caption{Coverage by domain.
             \llmvb{} matches or exceeds \panda{}~\rcff{} on six of seven domains; Robot is the exception, where every \pytrich{} configuration stops at the eleven smallest instances within the time limit.}
    \label{fig:app-rc2ff-coverage}
  \end{subfigure}
  \hfill
  \begin{subfigure}[t]{0.48\textwidth}
    \centering
    \includegraphics[width=\linewidth]{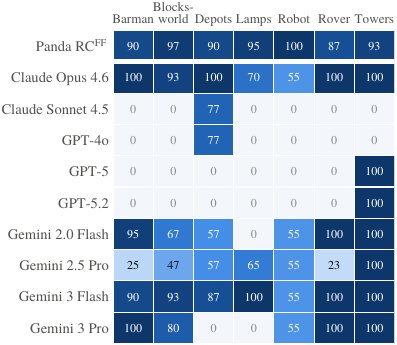}
    \caption{Coverage heatmap across models and domains.
             Claude Opus~4.6 and Gemini 3 Flash solve 148 instances each; GPT-4o, Claude Sonnet~4.5, GPT-5 and GPT-5.2 have results in one domain.
             Cells give the percentage of the instances in that domain solved by \llmvb{} or \rcff{}.}
    \label{fig:app-rc2ff-heatmap}
  \end{subfigure}
  \caption{Coverage results for \llmvb{} vs.\ \panda{}~\rcff{}: per-domain bar chart (a) and per-model heatmap (b).}
  \label{fig:app-rc2ff-coverage-group}
\end{figure}

\begin{figure}[ht]
  \centering
  \begin{subfigure}[t]{0.48\textwidth}
    \centering
    \includegraphics[width=\linewidth]{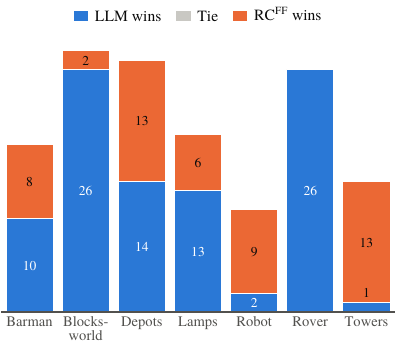}
    \caption{Win/loss on expanded nodes.
             \llmvb{} expands fewer nodes on 92 of 143 shared instances, every one in Rover and 26 of 28 in Blocksworld; \rcff{} wins Robot (9 of 11) and Towers (13 of 14), and Barman and Depots split about evenly.}
    \label{fig:app-rc2ff-winloss-nodes}
  \end{subfigure}
  \hfill
  \begin{subfigure}[t]{0.48\textwidth}
    \centering
    \includegraphics[width=\linewidth]{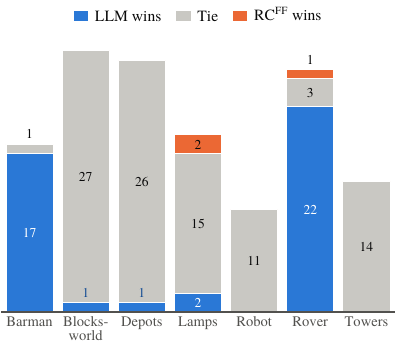}
    \caption{Win/loss on solution size.
             Shortest plan over all runs on each side: \llmvb{} is shorter on 43 of 143 shared instances, equal on 97 and longer on 3; the wins are Barman (17 of 18) and Rover (22 of 26), with one instance each in Blocksworld and Depots and two in Lamps.}
    \label{fig:app-rc2ff-winloss-size}
  \end{subfigure}
  \caption{Win/loss breakdown by domain: \llmvb{} vs.\ \panda{}~\rcff{} on expanded nodes (a) and solution size (b).
           Each bar segment counts shared instances where \llmvb{} expands fewer (win), equal (tie), or more (loss) nodes or actions.}
  \label{fig:app-rc2ff-winloss-group}
\end{figure}

\begin{figure}[ht]
  \centering
  \begin{subfigure}[t]{0.48\textwidth}
    \centering
    \includegraphics[width=\linewidth]{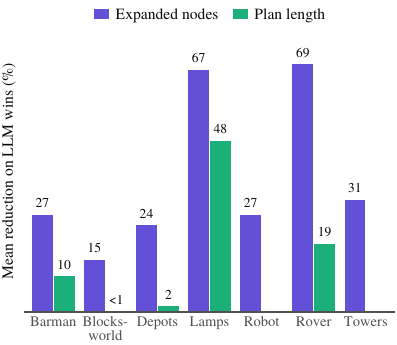}
    \caption{Average percentage improvement in expanded nodes on instances where \llmvb{} wins.
             Rover shows the largest mean reduction (69\%) and Lamps the next (67\%); Blocksworld the smallest (15\%).}
    \label{fig:app-rc2ff-improvement}
  \end{subfigure}
  \hfill
  \begin{subfigure}[t]{0.48\textwidth}
    \centering
    \includegraphics[width=\linewidth]{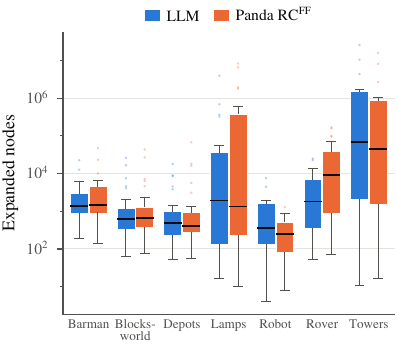}
    \caption{Boxplot of expanded nodes per domain on the problems solved by both systems, with the best run of each side on each instance.
             Rover shows the clearest \llmvb{} advantage; Robot and Towers favour \rcff{}.}
    \label{fig:app-rc2ff-boxplot}
  \end{subfigure}
  \caption{Search efficiency detail: \llmvb{} vs.\ \panda{}~\rcff{}.
           Panel (a) shows mean improvement on \llmvb{} wins; panel (b) shows the distribution of best node counts per domain on shared instances.}
  \label{fig:app-rc2ff-efficiency-group}
\end{figure}

\begin{figure}[ht]
  \centering
  \begin{subfigure}[t]{0.48\textwidth}
    \centering
    \includegraphics[width=\linewidth]{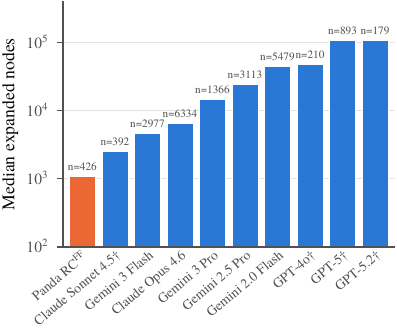}
    \caption{Median expanded nodes per model on problems each model solved.
             No LLM model matches the \panda{} medians (1{,}073 nodes for \rcff{} and 757 for \rclmcut{} over their solved runs, against 4{,}686 for Gemini 3 Flash and 6{,}599 for Claude Opus~4.6); the counts are planner-specific and are not comparable across the two systems. Models with results in only one domain are shown with a dagger.}
    \label{fig:app-rc2ff-models}
  \end{subfigure}
  \hfill
  \begin{subfigure}[t]{0.48\textwidth}
    \centering
    \includegraphics[width=\linewidth]{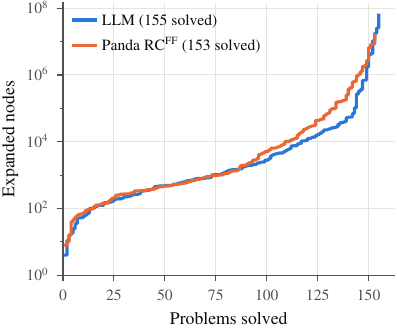}
    \caption{Cactus plot: problems solved within a given node-expansion budget.
             The two curves track each other up to a budget of ten thousand nodes and separate at the tail, where \llmvb{} solves the Rover and Barman instances that \rcff{} solves late or not at all; within ten thousand nodes \llmvb{} is ahead in Rover (20 against 13) and behind in Lamps and Towers by one.}
    \label{fig:app-rc2ff-cactus}
  \end{subfigure}
  \caption{Model ranking and cumulative coverage: \llmvb{} vs.\ \panda{}~\rcff{}.
           Panel (a) compares median search effort per LLM model against the two \panda{} variants; panel (b) shows how coverage accumulates as the node budget grows.}
  \label{fig:app-rc2ff-model-cactus-group}
\end{figure}

\FloatBarrier
\subsection{\panda{}~\rclmcut{}}
\label{app:panda-rc2lmc}

This subsection mirrors the \rcff{} comparison for completeness.
Since \panda{}~\rclmcut{} solves fewer problems than \panda{}~\rcff{}, the shared instance set is smaller in several domains.

\begin{figure}[ht]
  \centering
  \begin{subfigure}[t]{0.48\textwidth}
    \centering
    \includegraphics[width=\linewidth]{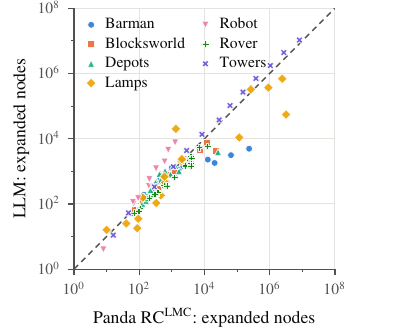}
    \caption{Expanded nodes: \llmvb{} vs.\ \panda{}~\rclmcut{}.
             \panda{}~\rclmcut{} has the lowest median node count of all baselines (757 over its solved runs), and \llmvb{} expands fewer nodes on 76 of the 125 shared instances, median ratio 0.88, winning every Rover instance and 25 of 26 in Blocksworld while losing Robot (10 of 11), Towers (12 of 13) and most of Depots.}
    \label{fig:app-rc2lmc-scatter-nodes}
  \end{subfigure}
  \hfill
  \begin{subfigure}[t]{0.48\textwidth}
    \centering
    \includegraphics[width=\linewidth]{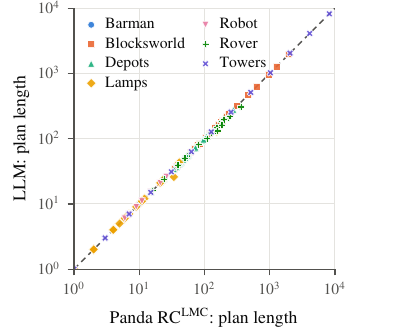}
    \caption{Solution size: \llmvb{} vs.\ \panda{}~\rclmcut{}.
             Points sit on the diagonal except in Barman and Rover: shortest plan over all runs, \llmvb{} is shorter on 18 of 125 shared instances, equal on 105 and longer on 2.}
    \label{fig:app-rc2lmc-scatter-size}
  \end{subfigure}
  \caption{Per-problem scatter plots: \llmvb{} vs.\ \panda{}~\rclmcut{} on expanded nodes (a) and solution size (b).
           Points below the diagonal indicate \llmvb{} advantage.}
  \label{fig:app-rc2lmc-scatter-group}
\end{figure}

\begin{figure}[ht]
  \centering
  \begin{subfigure}[t]{0.48\textwidth}
    \centering
    \includegraphics[width=\linewidth]{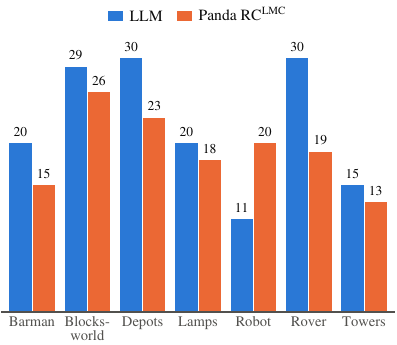}
    \caption{Coverage by domain.
             \panda{}~\rclmcut{} solves 134 of 180 problems, against 155 for \llmvb{}; the gap is largest in Rover (19 against 30) and Depots (23 against 30).}
    \label{fig:app-rc2lmc-coverage}
  \end{subfigure}
  \hfill
  \begin{subfigure}[t]{0.48\textwidth}
    \centering
    \includegraphics[width=\linewidth]{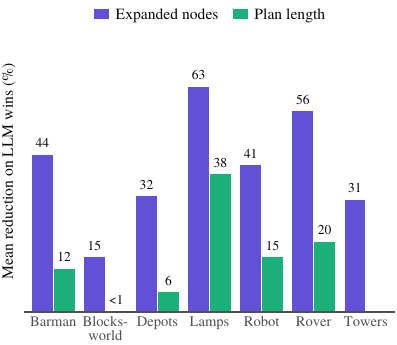}
    \caption{Average percentage improvement in expanded nodes on instances where \llmvb{} wins.
             On the instances \llmvb{} wins, the mean reduction is largest in Lamps (63\%) and Barman (44\%) and smallest in Blocksworld (12\%).}
    \label{fig:app-rc2lmc-improvement}
  \end{subfigure}
  \caption{Coverage and search-efficiency summary: \llmvb{} vs.\ \panda{}~\rclmcut{}.}
  \label{fig:app-rc2lmc-summary-group}
\end{figure}

\FloatBarrier
\subsection{Plan Length by Domain}
\label{app:planlength}

Table~\ref{tab:planlength-intersection} reports median plan length restricted to the problems solved by all seven systems, with each entry being the virtual best over the three search algorithms.
Every system returns the same median length in Blocksworld, Depots, Robot, Towers and Lamps.
The systems differ only in Barman and Rover: in Barman \rcff{} returns plans six actions longer than the LLM heuristics in the median, and in Rover \rcff{} and \pandalmcount{} return plans 11 and 56 actions longer.
\citet{YousefiSchmautzHaslumEtAl2025} establish that A$^*$ is incomplete for total-order \htn{} planning, so differences in plan length across planners with different internal search strategies are unsurprising (see also Section~\ref{sec:planquality}).

\begin{table}[t]
  \caption{Median plan length (primitive actions) per domain, restricted to the problems solved by all seven systems. Each entry is the virtual best over the three search algorithms (GBFS, A*, WA*).}
  \label{tab:planlength-intersection}
  \centering
  \begin{tabular}{lrrrrrrrr}
    \toprule
    & & \multicolumn{3}{c}{\panda{}} & \multicolumn{4}{c}{\pytrich{}} \\
    \cmidrule(lr){3-5}\cmidrule(lr){6-9}
    Domain & $n$ & \rcff{} & \rclmcut{} & \pandalmcount{} & \tdg{} & \lmcount{} & \coiv{} & \llmvb{} \\
    \midrule
    Barman-BDI             & 15      & 44     & 38       & 38       & 43      & 40        & 38      & 38 \\
    Blocksworld-GTOHP      & 20      & 146    & 146      & 146      & 146     & 146       & 146     & 146 \\
    Depots                 & 23      & 57     & 57       & 57       & 57      & 57        & 57      & 57 \\
    Robot                  & 11      & 11     & 11       & 11       & 11      & 11        & 11      & 11 \\
    Rover-GTOHP            & 16      & 101    & 96       & 146      & 92      & 90        & 90      & 90 \\
    Towers                 & 13      & 127    & 127      & 127      & 127     & 127       & 127     & 127 \\
    Lamps                  & 12      & 4      & 4        & 4        & 4       & 4         & 4       & 4 \\
    \midrule
    Median                 &         & 57     & 57       & 57       & 57      & 57        & 57      & 57 \\
    \bottomrule
  \end{tabular}
\end{table}

\FloatBarrier
\subsection{Median Expanded Nodes by Model and Algorithm}
\label{app:median-nodes}

Table~\ref{tab:median_nodes} reports median expanded nodes broken down by model, algorithm, and domain.
Dashes indicate that the model produced no working heuristics for that domain.
Claude Opus~4.6 rows are highlighted in bold; the overall median pools all problems solved by each model under each algorithm.

\begin{table}[t]
  \caption{Median expanded nodes over solved runs by model, algorithm and domain. \llmvb{} pools the five main models; the \pytrich{} baseline rows pool their runs under the three algorithms, which the record does not label.}
  \label{tab:median_nodes}
  \centering
  \footnotesize
  \setlength{\tabcolsep}{3pt}
  \begin{tabular}{llrrrrrrrr}
    \toprule
    \textbf{Model} & \textbf{Algo} & \textbf{Barman} & \textbf{Blocks.} & \textbf{Depots} & \textbf{Robot} & \textbf{Rover} & \textbf{Towers} & \textbf{Lamps} & \textbf{Overall} \\
    \midrule
  \llmvb{} & GBFS & 47,074 & 1,358 & 3,861 & 1,419 & 13,785 & 110,252 & 4,229 & 8,761 \\
  \llmvb{} & WA* & 76,541 & 1,904 & 6,571 & 1,495 & 15,570 & 110,208 & 8,298 & 11,934 \\
  \llmvb{} & A* & 377,048 & 25,352 & 23,502 & 1,495 & 16,425 & 110,375 & 7,260 & 21,602 \\
    \midrule
  GPT-4o & GBFS & \multicolumn{1}{c}{---} & \multicolumn{1}{c}{---} & 8,614 & \multicolumn{1}{c}{---} & \multicolumn{1}{c}{---} & \multicolumn{1}{c}{---} & \multicolumn{1}{c}{---} & 8,614 \\
   & WA* & \multicolumn{1}{c}{---} & \multicolumn{1}{c}{---} & 74,097 & \multicolumn{1}{c}{---} & \multicolumn{1}{c}{---} & \multicolumn{1}{c}{---} & \multicolumn{1}{c}{---} & 74,097 \\
   & A* & \multicolumn{1}{c}{---} & \multicolumn{1}{c}{---} & 74,097 & \multicolumn{1}{c}{---} & \multicolumn{1}{c}{---} & \multicolumn{1}{c}{---} & \multicolumn{1}{c}{---} & 74,097 \\
  \textbf{\coiv{}} & \textbf{GBFS} & \textbf{15,523} & \textbf{540} & \textbf{953} & \textbf{1,185} & \textbf{29,136} & \textbf{109,274} & \textbf{10,982} & \textbf{3,273} \\
  \textbf{} & \textbf{WA*} & \textbf{26,788} & \textbf{647} & \textbf{1,167} & \textbf{1,117} & \textbf{36,430} & \textbf{109,378} & \textbf{10,306} & \textbf{4,972} \\
  \textbf{} & \textbf{A*} & \textbf{515,107} & \textbf{47,166} & \textbf{7,776} & \textbf{1,116} & \textbf{17,805} & \textbf{109,484} & \textbf{8,319} & \textbf{20,835} \\
  Claude Sonnet~4.5 & GBFS & \multicolumn{1}{c}{---} & \multicolumn{1}{c}{---} & 1,128 & \multicolumn{1}{c}{---} & \multicolumn{1}{c}{---} & \multicolumn{1}{c}{---} & \multicolumn{1}{c}{---} & 1,128 \\
   & WA* & \multicolumn{1}{c}{---} & \multicolumn{1}{c}{---} & 1,860 & \multicolumn{1}{c}{---} & \multicolumn{1}{c}{---} & \multicolumn{1}{c}{---} & \multicolumn{1}{c}{---} & 1,860 \\
   & A* & \multicolumn{1}{c}{---} & \multicolumn{1}{c}{---} & 8,436 & \multicolumn{1}{c}{---} & \multicolumn{1}{c}{---} & \multicolumn{1}{c}{---} & \multicolumn{1}{c}{---} & 8,436 \\
  GPT-5 & GBFS & \multicolumn{1}{c}{---} & \multicolumn{1}{c}{---} & \multicolumn{1}{c}{---} & \multicolumn{1}{c}{---} & \multicolumn{1}{c}{---} & 109,274 & \multicolumn{1}{c}{---} & 109,274 \\
   & WA* & \multicolumn{1}{c}{---} & \multicolumn{1}{c}{---} & \multicolumn{1}{c}{---} & \multicolumn{1}{c}{---} & \multicolumn{1}{c}{---} & 109,378 & \multicolumn{1}{c}{---} & 109,378 \\
   & A* & \multicolumn{1}{c}{---} & \multicolumn{1}{c}{---} & \multicolumn{1}{c}{---} & \multicolumn{1}{c}{---} & \multicolumn{1}{c}{---} & 109,484 & \multicolumn{1}{c}{---} & 109,484 \\
  GPT-5.2 & GBFS & \multicolumn{1}{c}{---} & \multicolumn{1}{c}{---} & \multicolumn{1}{c}{---} & \multicolumn{1}{c}{---} & \multicolumn{1}{c}{---} & 109,274 & \multicolumn{1}{c}{---} & 109,274 \\
   & WA* & \multicolumn{1}{c}{---} & \multicolumn{1}{c}{---} & \multicolumn{1}{c}{---} & \multicolumn{1}{c}{---} & \multicolumn{1}{c}{---} & 109,378 & \multicolumn{1}{c}{---} & 109,378 \\
   & A* & \multicolumn{1}{c}{---} & \multicolumn{1}{c}{---} & \multicolumn{1}{c}{---} & \multicolumn{1}{c}{---} & \multicolumn{1}{c}{---} & 109,484 & \multicolumn{1}{c}{---} & 109,484 \\
  Gemini 2.0 Flash & GBFS & 909,166 & 39,582 & 18,379 & 1,498 & 80,128 & 111,058 & \multicolumn{1}{c}{---} & 20,795 \\
   & WA* & 3,244,057 & 64,696 & 74,097 & 1,582 & 71,108 & 111,162 & \multicolumn{1}{c}{---} & 48,631 \\
   & A* & 3,244,057 & 66,868 & 74,097 & 1,582 & 71,108 & 111,162 & \multicolumn{1}{c}{---} & 48,631 \\
  Gemini 2.5 Pro & GBFS & 3,376,203 & 62,760 & 18,379 & 1,498 & 80,128 & 111,058 & 4,229 & 31,074 \\
   & WA* & 3,244,057 & 69,069 & 74,097 & 1,582 & 71,108 & 111,162 & 11,527 & 32,970 \\
   & A* & 3,244,057 & 69,069 & 74,097 & 1,582 & 71,108 & 111,162 & 8,319 & 23,502 \\
  Gemini 3 Flash & GBFS & 80,200 & 1,511 & 1,102 & 797 & 5,415 & 106,154 & 3,253 & 3,765 \\
   & WA* & 71,556 & 1,702 & 1,113 & 788 & 5,451 & 105,722 & 2,479 & 3,942 \\
   & A* & 110,384 & 4,522 & 4,464 & 789 & 15,418 & 108,112 & 5,035 & 6,554 \\
  Gemini 3 Pro & GBFS & 10,558 & 1,357 & \multicolumn{1}{c}{---} & 877 & 2,624 & 110,638 & \multicolumn{1}{c}{---} & 14,639 \\
   & WA* & 12,714 & 1,519 & \multicolumn{1}{c}{---} & 884 & 2,720 & 110,468 & \multicolumn{1}{c}{---} & 14,705 \\
   & A* & 15,784 & 3,933 & \multicolumn{1}{c}{---} & 957 & 4,224 & 110,671 & \multicolumn{1}{c}{---} & 23,425 \\
    \midrule
  \pandalmcount{} & GBFS & 1,596 & 789 & 412 & 596 & 19,535 & 44,468 & 2,365 & 1,196 \\
   & WA* & 16,604 & 984 & 482 & 621 & 61,166 & 44,468 & 1,226 & 1,378 \\
   & A* & 357,790 & 2,667 & 14,232 & 158 & 126,745 & 44,468 & 1,289 & 3,597 \\
  \panda{}~\rcadd{} & GBFS & 1,438 & 692 & 456 & 1,060 & 2,766 & 44,468 & 1,014 & 1,010 \\
   & WA* & 1,546 & 692 & 506 & 1,076 & 2,807 & 44,468 & 1,010 & 1,074 \\
   & A* & 1,452 & 692 & 691 & 825 & 3,612 & 44,468 & 643 & 1,090 \\
  \panda{}~\rcff{} & GBFS & 1,438 & 682 & 398 & 1,067 & 10,106 & 44,468 & 1,346 & 1,024 \\
   & WA* & 1,992 & 737 & 482 & 966 & 11,394 & 44,468 & 2,300 & 1,079 \\
   & A* & 3,169 & 636 & 870 & 361 & 21,919 & 44,468 & 671 & 1,107 \\
  \panda{}~\rclmcut{} & GBFS & 1,173 & 629 & 334 & 1,019 & 2,595 & 28,236 & 850 & 721 \\
   & WA* & 1,947 & 655 & 342 & 622 & 2,660 & 28,236 & 789 & 727 \\
   & A* & 7,910 & 618 & 6,106 & 392 & 701 & 28,236 & 648 & 1,005 \\
  \lmcount{} & all & 24,944 & 1,457 & 1,414 & 1,450 & 17,389 & 111,162 & 6,462 & 4,936 \\
  \tdg{} & all & 84,196 & 1,468 & 3,297 & 891 & 3,512 & 109,972 & 1,512 & 4,730 \\
    \bottomrule
  \end{tabular}
\end{table}

\FloatBarrier
\subsection{\pytrich{}-\tdg{}}
\label{app:pytrich-tdg}

This subsection compares \llmvb{} against \pytrich{} guided by \tdg{}.
The figures report the same metrics as above: coverage, expanded nodes, and plan length on shared solved instances.

\paragraph{Why the margin over \tdg{} varies by domain.}
\tdg{} gives every method a weight of one and every operator its cost, estimates each pending task by its cheapest method, and sums over the task network, so it never reads the state.
Wherever a task has a method with no subtasks whose precondition is that the task is already achieved, \tdg{} assumes that method, and its estimate reduces to the number of pending tasks whatever the state.
In Barman every \texttt{Achieve} task has such a method, for instance \texttt{AddIngredientToShotNull} when the shot already contains the ingredient, while the true cost depends on whether the shot is clean, the shaker empty and a hand free; \tdg{} expands 1,460 nodes per plan action at the median against 29 for the best Claude Opus~4.6 heuristic, and the LLM virtual best expands 98\% fewer nodes.
Blocksworld and Depots have the same shape, with \texttt{do\_put\_on} and \texttt{do\_clear} decomposing to \texttt{nop} when the block or crate is already in place, while the true cost is the height of the stack above it; the LLM virtual best expands 55\% and 59\% fewer nodes.
In Lamps \texttt{play} and each \texttt{propagate} task end with an empty method, while the cost is the number of lamps in the row that still share the flipped status; 76\% fewer nodes.
In Rover \texttt{get\_soil\_data} decomposes into navigate, empty the store, sample and send, each with a fixed cheapest cost, so \tdg{} estimates about four actions per sample goal and six per image goal and only the navigation distance depends on the state; \tdg{} and the best LLM heuristic both expand about 12 nodes per plan action, and the two are at parity.
In Towers the preconditions of the methods are mutually exclusive, \texttt{on ?r ?t1} against \texttt{on ?r ?r1} and \texttt{exchangeLR} against \texttt{exchangeRL}, so at most one method applies in any state; every system, \panda{} included, expands the decomposition tree of about four hundred nodes per action and returns the same plan, and the heuristic cannot matter.
Robot is the exception.
\texttt{achieve-goals} has the empty method \texttt{finished}, so \tdg{} is as uninformative as in Barman, yet the LLM heuristics expand only 23\% fewer nodes and no \pytrich{} heuristic solves more than the eleven smallest instances; the recursive \texttt{achieve-goals}, \texttt{release} and \texttt{move\_abstract} tasks place every state one decomposition step from any other, so the difficulty lies in the branching rather than in the estimate.
Table~\ref{tab:domain-characteristics} gives the structure of each lifted domain beside the median node ratio of \llmvb{} against \tdg{}, ordered by that ratio.
The margin tracks the decomposition branching factor only loosely.
Rover and Towers, the two domains with the fewest methods per task, are the two where \tdg{} is at parity, and the domains with two or more methods per task show the largest gains, but what separates Barman and Lamps from Blocksworld and Depots is not in the table.
It is which of those methods is the empty one, as the paragraph above describes.

\begin{table}[htbp]
  \centering
  \caption{Structure of the seven domains, measured on the lifted \hddl{} domain file, and the median node ratio of \llmvb{} against \tdg{} on shared instances.
           $|A|$, $|C|$ and $|M|$ count primitive action, compound task and method schemas; methods per task and subtasks per method are mean (max); depth is the longest chain in the condensation of the task decomposition graph.}
  \label{tab:domain-characteristics}
  \footnotesize
  \setlength{\tabcolsep}{4pt}
  \begin{tabular}{lrrrrrrlr}
\toprule
Domain & $|A|$ & $|C|$ & $|M|$ & Methods/task & Subtasks/method & Depth & Recursive & Node ratio vs \tdg{} \\
\midrule
Barman & 11 & 10 & 22 & 2.20 (3) & 1.82 (6) & 6 & no & 0.02 \\
Lamps & 1 & 6 & 15 & 2.50 (3) & 1.00 (5) & 2 & yes & 0.24 \\
Depots & 6 & 6 & 12 & 2.00 (4) & 2.42 (4) & 3 & yes & 0.41 \\
Blocksworld & 5 & 4 & 8 & 2.00 (2) & 2.00 (4) & 1 & yes & 0.45 \\
Robot & 4 & 6 & 11 & 1.83 (4) & 1.45 (2) & 1 & yes & 0.77 \\
Towers & 1 & 5 & 8 & 1.60 (3) & 1.25 (2) & 3 & yes & 0.96 \\
Rover & 14 & 10 & 16 & 1.60 (3) & 2.12 (4) & 3 & yes & 1.00 \\
\bottomrule
\end{tabular}

\end{table}

\begin{figure}[ht]
  \centering
  \begin{subfigure}[t]{0.48\textwidth}
    \centering
    \includegraphics[width=\linewidth]{./new-plots/scatter_expanded_nodes_tdg_v2}
    \caption{Expanded nodes: \llmvb{} vs.\ \tdg{}.
             \llmvb{} expands fewer nodes on 107 of the 132 shared instances, ties on 24, most of them in Rover, and loses one, in Lamps.
             The 48 instances \tdg{} does not solve are absent from this plot.}
    \label{fig:app-tdg-scatter-nodes}
  \end{subfigure}
  \hfill
  \begin{subfigure}[t]{0.48\textwidth}
    \centering
    \includegraphics[width=\linewidth]{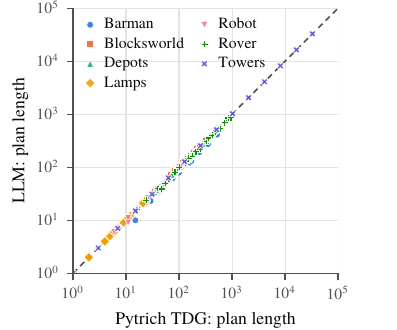}
    \caption{Solution size: \llmvb{} vs.\ \tdg{}.
             Shortest plan over all runs on each side: \llmvb{} is shorter on 43 of 132 shared instances and equal on the rest, with the wins in Barman (all 18), Rover (23 of 30) and two Robot instances.}
    \label{fig:app-tdg-scatter-size}
  \end{subfigure}
  \caption{Per-problem scatter plots: \llmvb{} vs.\ \tdg{} on expanded nodes (a) and solution size (b).
           Points below the diagonal indicate \llmvb{} advantage.}
  \label{fig:app-tdg-scatter-group}
\end{figure}

\begin{figure}[ht]
  \centering
  \begin{subfigure}[t]{0.48\textwidth}
    \centering
    \includegraphics[width=\linewidth]{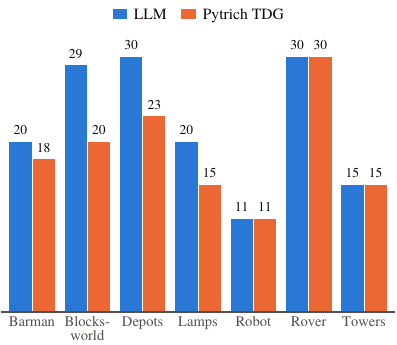}
    \caption{Coverage by domain.
             \llmvb{} solves 155 problems and \tdg{} 132, and \llmvb{} solves every instance that \tdg{} solves.}
    \label{fig:app-tdg-coverage}
  \end{subfigure}
  \hfill
  \begin{subfigure}[t]{0.48\textwidth}
    \centering
    \includegraphics[width=\linewidth]{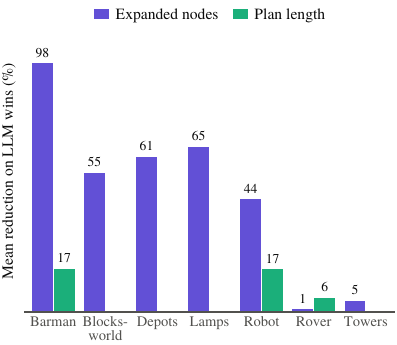}
    \caption{Average percentage improvement in expanded nodes on shared instances where \llmvb{} wins.
             Barman shows the largest mean reduction (98\%) and Lamps the next (65\%); Towers (5\%) and Rover (1\%) the smallest, where the method library fixes most of the cost.}
    \label{fig:app-tdg-improvement}
  \end{subfigure}
  \caption{Coverage and search-efficiency summary: \llmvb{} vs.\ \tdg{}.}
  \label{fig:app-tdg-summary-group}
\end{figure}

\FloatBarrier
\subsection{\pytrich{}-\lmcount{}}
\label{app:pytrich-lmcount}

This subsection compares \llmvb{} against \pytrich{} guided by \lmcount{}, the bidirectional landmark-count heuristic.

\begin{figure}[ht]
  \centering
  \begin{subfigure}[t]{0.48\textwidth}
    \centering
    \includegraphics[width=\linewidth]{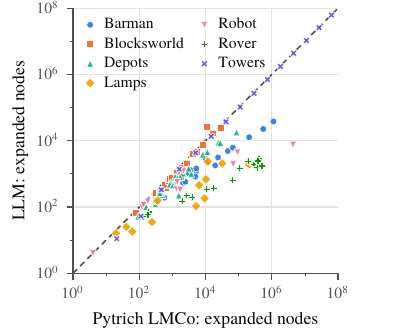}
    \caption{Expanded nodes: \llmvb{} vs.\ \lmcount{}.
             \llmvb{} expands fewer nodes on 126 of the 129 shared instances, median ratio 0.46, and loses two, in Blocksworld; every domain is won outright except Blocksworld (27 of 29).
             Points below the diagonal indicate an \llmvb{} advantage.}
    \label{fig:app-lmcount-scatter-nodes}
  \end{subfigure}
  \hfill
  \begin{subfigure}[t]{0.48\textwidth}
    \centering
    \includegraphics[width=\linewidth]{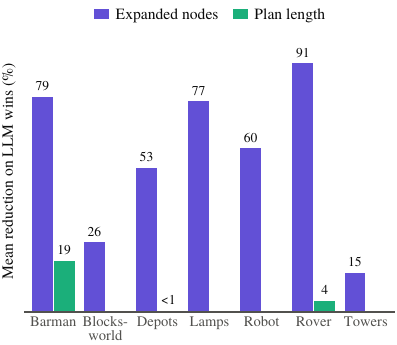}
    \caption{Average percentage improvement in expanded nodes on shared instances where \llmvb{} wins.
             The median node ratio per domain is 0.02 in Rover, 0.11 in Lamps, 0.22 in Barman, 0.41 in Robot, 0.47 in Depots, 0.78 in Blocksworld and 0.92 in Towers; plan lengths tie on every shared instance outside Barman and Rover.}
    \label{fig:app-lmcount-improvement}
  \end{subfigure}
  \caption{Search efficiency within \pytrich{}: \llmvb{} vs.\ \lmcount{} on per-problem expanded nodes (a) and average improvement per domain (b).}
  \label{fig:app-lmcount-group}
\end{figure}
\FloatBarrier

\section{Heuristic API Details}
\label{app:api}

This appendix describes the implementation-level interface used for all LLM-generated heuristics.

\textbf{Heuristic class.}
Each heuristic subclasses \texttt{Heuristic} and implements two methods:
\begin{itemize}
  \item \texttt{initialize(self, model: Model, initial\_node: HTNNode)}: called once before search for preprocessing.
  \item \texttt{\_\_call\_\_(self, parent\_node: HTNNode, node: HTNNode) -> int}: called at each node expansion; returns a non-negative integer estimate of remaining solution cost.
\end{itemize}

\textbf{Model interface.}
The \texttt{Model} object exposes grounded planning objects including
\texttt{facts}, \texttt{operators}, \texttt{abstract\_tasks}, \texttt{decompositions}, and the goal bitset \texttt{goals}.

\textbf{Node interface.}
Each \texttt{HTNNode} provides:
\begin{itemize}
  \item \texttt{node.state}: integer bitset encoding the current state;
  \item \texttt{node.task\_network}: ordered remaining tasks (each tagged primitive or abstract);
  \item \texttt{node.decomposition}: the method applied to reach this node, or \texttt{None} at the root and after a primitive action.
\end{itemize}

\textbf{Bitset access.}
Fact truth checks use constant-time bit operations,
\texttt{(node.state >{}> fact.global\_id) \& 1}, and operator preconditions/effects are also represented as bitsets.
This allows efficient reachability-style computations without string parsing.

\textbf{Grounded fact naming.}
Grounded facts follow runtime names such as \texttt{+predicate[arg1,arg2]} and \texttt{-predicate[arg1,arg2]}, which differ from raw \hddl{} syntax.
Prompts include this convention explicitly to avoid generation errors.

\clearpage
\section*{NeurIPS Paper Checklist}

\begin{enumerate}

\item {\bf Claims}
    \item[] Question: Do the main claims made in the abstract and introduction accurately reflect the paper's contributions and scope?
    \item[] Answer: \answerYes{}
    \item[] Justification: The abstract and Section~\ref{sec:introduction} state three contributions, the pipeline in \pytrich{}, the evaluation of nine models on seven domains against domain-independent heuristics and \panda{}, and the analysis of search effort, domain structure, models and algorithms, each supported by Section~\ref{sec:results} and Section~\ref{sec:discussion}.
    \item[] Guidelines:
    \begin{itemize}
        \item The answer \answerNA{} means that the abstract and introduction do not include the claims made in the paper.
        \item The abstract and/or introduction should clearly state the claims made, including the contributions made in the paper and important assumptions and limitations. A \answerNo{} or \answerNA{} answer to this question will not be perceived well by the reviewers. 
        \item The claims made should match theoretical and experimental results, and reflect how much the results can be expected to generalize to other settings. 
        \item It is fine to include aspirational goals as motivation as long as it is clear that these goals are not attained by the paper. 
    \end{itemize}

\item {\bf Limitations}
    \item[] Question: Does the paper discuss the limitations of the work performed by the authors?
    \item[] Answer: \answerYes{}
    \item[] Justification: Section~\ref{sec:conclusion} states four limitations: the evaluation covers seven total-order \htn{} domains and leaves partial-order, numerical and temporal planning open; the virtual best is an upper bound and the fixed configuration's single selection instance filters without ranking; generation is stochastic and the prompts contain two evaluated instances per domain, which Appendix~\ref{app:prompt-context} recomputes without; and the heuristics cannot run inside \panda{}, so that comparison rests on coverage.
    \item[] Guidelines:
    \begin{itemize}
        \item The answer \answerNA{} means that the paper has no limitation while the answer \answerNo{} means that the paper has limitations, but those are not discussed in the paper. 
        \item The authors are encouraged to create a separate ``Limitations'' section in their paper.
        \item The paper should point out any strong assumptions and how robust the results are to violations of these assumptions (e.g., independence assumptions, noiseless settings, model well-specification, asymptotic approximations only holding locally). The authors should reflect on how these assumptions might be violated in practice and what the implications would be.
        \item The authors should reflect on the scope of the claims made, e.g., if the approach was only tested on a few datasets or with a few runs. In general, empirical results often depend on implicit assumptions, which should be articulated.
        \item The authors should reflect on the factors that influence the performance of the approach. For example, a facial recognition algorithm may perform poorly when image resolution is low or images are taken in low lighting. Or a speech-to-text system might not be used reliably to provide closed captions for online lectures because it fails to handle technical jargon.
        \item The authors should discuss the computational efficiency of the proposed algorithms and how they scale with dataset size.
        \item If applicable, the authors should discuss possible limitations of their approach to address problems of privacy and fairness.
        \item While the authors might fear that complete honesty about limitations might be used by reviewers as grounds for rejection, a worse outcome might be that reviewers discover limitations that aren't acknowledged in the paper. The authors should use their best judgment and recognize that individual actions in favor of transparency play an important role in developing norms that preserve the integrity of the community. Reviewers will be specifically instructed to not penalize honesty concerning limitations.
    \end{itemize}

\item {\bf Theory assumptions and proofs}
    \item[] Question: For each theoretical result, does the paper provide the full set of assumptions and a complete (and correct) proof?
    \item[] Answer: \answerNA{}
    \item[] Justification: The paper presents no theoretical results; all contributions are empirical.
    \item[] Guidelines:
    \begin{itemize}
        \item The answer \answerNA{} means that the paper does not include theoretical results. 
        \item All the theorems, formulas, and proofs in the paper should be numbered and cross-referenced.
        \item All assumptions should be clearly stated or referenced in the statement of any theorems.
        \item The proofs can either appear in the main paper or the supplemental material, but if they appear in the supplemental material, the authors are encouraged to provide a short proof sketch to provide intuition. 
        \item Inversely, any informal proof provided in the core of the paper should be complemented by formal proofs provided in appendix or supplemental material.
        \item Theorems and Lemmas that the proof relies upon should be properly referenced. 
    \end{itemize}

    \item {\bf Experimental result reproducibility}
    \item[] Question: Does the paper fully disclose all the information needed to reproduce the main experimental results of the paper to the extent that it affects the main claims and/or conclusions of the paper (regardless of whether the code and data are provided or not)?
    \item[] Answer: \answerYes{}
    \item[] Justification: Section~\ref{sec:method} describes the generate-evaluate-select pipeline; Appendix~\ref{app:prompts} reproduces the full prompt text; Appendix~\ref{app:api} documents the heuristic interface; and Appendix~\ref{app:expdetails} provides model versions, generation parameters, and resource limits. Both baseline planners are publicly available.
    \item[] Guidelines:
    \begin{itemize}
        \item The answer \answerNA{} means that the paper does not include experiments.
        \item If the paper includes experiments, a \answerNo{} answer to this question will not be perceived well by the reviewers: Making the paper reproducible is important, regardless of whether the code and data are provided or not.
        \item If the contribution is a dataset and\slash or model, the authors should describe the steps taken to make their results reproducible or verifiable. 
        \item Depending on the contribution, reproducibility can be accomplished in various ways. For example, if the contribution is a novel architecture, describing the architecture fully might suffice, or if the contribution is a specific model and empirical evaluation, it may be necessary to either make it possible for others to replicate the model with the same dataset, or provide access to the model. In general. releasing code and data is often one good way to accomplish this, but reproducibility can also be provided via detailed instructions for how to replicate the results, access to a hosted model (e.g., in the case of a large language model), releasing of a model checkpoint, or other means that are appropriate to the research performed.
        \item While NeurIPS does not require releasing code, the conference does require all submissions to provide some reasonable avenue for reproducibility, which may depend on the nature of the contribution. For example
        \begin{enumerate}
            \item If the contribution is primarily a new algorithm, the paper should make it clear how to reproduce that algorithm.
            \item If the contribution is primarily a new model architecture, the paper should describe the architecture clearly and fully.
            \item If the contribution is a new model (e.g., a large language model), then there should either be a way to access this model for reproducing the results or a way to reproduce the model (e.g., with an open-source dataset or instructions for how to construct the dataset).
            \item We recognize that reproducibility may be tricky in some cases, in which case authors are welcome to describe the particular way they provide for reproducibility. In the case of closed-source models, it may be that access to the model is limited in some way (e.g., to registered users), but it should be possible for other researchers to have some path to reproducing or verifying the results.
        \end{enumerate}
    \end{itemize}

\item {\bf Open access to data and code}
    \item[] Question: Does the paper provide open access to the data and code, with sufficient instructions to faithfully reproduce the main experimental results, as described in supplemental material?
    \item[] Answer: \answerYes{}
    \item[] Justification: The code, the generated heuristics, the result files and the scripts that produce every table and figure are released as a repository archived on Zenodo~\citep{MeneguzziBuchweitzCorreaEtAl2026Software}, with a README that gives the exact commands to rebuild the \panda{} tools at pinned commits, ground the benchmarks, submit the job arrays and compare the outcome with the released result files. The IPC benchmark domains are publicly available, and the full prompt and heuristic API are reproduced in Appendix~\ref{app:prompts} and Appendix~\ref{app:api}.
    \item[] Guidelines:
    \begin{itemize}
        \item The answer \answerNA{} means that paper does not include experiments requiring code.
        \item Please see the NeurIPS code and data submission guidelines (\url{https://neurips.cc/public/guides/CodeSubmissionPolicy}) for more details.
        \item While we encourage the release of code and data, we understand that this might not be possible, so \answerNo{} is an acceptable answer. Papers cannot be rejected simply for not including code, unless this is central to the contribution (e.g., for a new open-source benchmark).
        \item The instructions should contain the exact command and environment needed to run to reproduce the results. See the NeurIPS code and data submission guidelines (\url{https://neurips.cc/public/guides/CodeSubmissionPolicy}) for more details.
        \item The authors should provide instructions on data access and preparation, including how to access the raw data, preprocessed data, intermediate data, and generated data, etc.
        \item The authors should provide scripts to reproduce all experimental results for the new proposed method and baselines. If only a subset of experiments are reproducible, they should state which ones are omitted from the script and why.
        \item At submission time, to preserve anonymity, the authors should release anonymized versions (if applicable).
        \item Providing as much information as possible in supplemental material (appended to the paper) is recommended, but including URLs to data and code is permitted.
    \end{itemize}

\item {\bf Experimental setting/details}
    \item[] Question: Does the paper specify all the training and test details (e.g., data splits, hyperparameters, how they were chosen, type of optimizer) necessary to understand the results?
    \item[] Answer: \answerYes{}
    \item[] Justification: Section~\ref{sec:setup} specifies the benchmark domains, LLM models, search algorithms, resource limits (1 CPU core, 8~GB RAM, 30-minute wall time), and evaluation metrics. Appendix~\ref{app:expdetails} provides model versions, generation parameters, and further infrastructure details.
    \item[] Guidelines:
    \begin{itemize}
        \item The answer \answerNA{} means that the paper does not include experiments.
        \item The experimental setting should be presented in the core of the paper to a level of detail that is necessary to appreciate the results and make sense of them.
        \item The full details can be provided either with the code, in appendix, or as supplemental material.
    \end{itemize}

\item {\bf Experiment statistical significance}
    \item[] Question: Does the paper report error bars suitably and correctly defined or other appropriate information about the statistical significance of the experiments?
    \item[] Answer: \answerNA{}
    \item[] Justification: Every planner run is deterministic given the heuristic, the instance and the search algorithm, so coverage and expanded nodes carry no sampling error at run time; the stochastic step is candidate generation, which Section~\ref{sec:conclusion} lists as a limitation, and Appendix~\ref{app:ablation} reports that two draws of twenty candidates from the same prompt can differ by a factor of 2.7 in median nodes. Error bars over repeated generation would require regenerating every model and domain and are not reported.
    \item[] Guidelines:
    \begin{itemize}
        \item The answer \answerNA{} means that the paper does not include experiments.
        \item The authors should answer \answerYes{} if the results are accompanied by error bars, confidence intervals, or statistical significance tests, at least for the experiments that support the main claims of the paper.
        \item The factors of variability that the error bars are capturing should be clearly stated (for example, train/test split, initialization, random drawing of some parameter, or overall run with given experimental conditions).
        \item The method for calculating the error bars should be explained (closed form formula, call to a library function, bootstrap, etc.)
        \item The assumptions made should be given (e.g., Normally distributed errors).
        \item It should be clear whether the error bar is the standard deviation or the standard error of the mean.
        \item It is OK to report 1-sigma error bars, but one should state it. The authors should preferably report a 2-sigma error bar than state that they have a 96\% CI, if the hypothesis of Normality of errors is not verified.
        \item For asymmetric distributions, the authors should be careful not to show in tables or figures symmetric error bars that would yield results that are out of range (e.g., negative error rates).
        \item If error bars are reported in tables or plots, the authors should explain in the text how they were calculated and reference the corresponding figures or tables in the text.
    \end{itemize}

\item {\bf Experiments compute resources}
    \item[] Question: For each experiment, does the paper provide sufficient information on the computer resources (type of compute workers, memory, time of execution) needed to reproduce the experiments?
    \item[] Answer: \answerYes{}
    \item[] Justification: Section~\ref{sec:setup} states the per-problem resource allocation (1 CPU core, 8~GB RAM, 30-minute wall-clock limit). Appendix~\ref{app:expdetails} provides cluster details and additional execution information for both \pytrich{} and \panda{} runs.
    \item[] Guidelines:
    \begin{itemize}
        \item The answer \answerNA{} means that the paper does not include experiments.
        \item The paper should indicate the type of compute workers CPU or GPU, internal cluster, or cloud provider, including relevant memory and storage.
        \item The paper should provide the amount of compute required for each of the individual experimental runs as well as estimate the total compute. 
        \item The paper should disclose whether the full research project required more compute than the experiments reported in the paper (e.g., preliminary or failed experiments that didn't make it into the paper). 
    \end{itemize}
    
\item {\bf Code of ethics}
    \item[] Question: Does the research conducted in the paper conform, in every respect, with the NeurIPS Code of Ethics \url{https://neurips.cc/public/EthicsGuidelines}?
    \item[] Answer: \answerYes{}
    \item[] Justification: The research uses publicly available LLM APIs and IPC benchmark domains, involves no human subjects or sensitive data, and raises no ethical concerns identified in the NeurIPS Code of Ethics.
    \item[] Guidelines:
    \begin{itemize}
        \item The answer \answerNA{} means that the authors have not reviewed the NeurIPS Code of Ethics.
        \item If the authors answer \answerNo, they should explain the special circumstances that require a deviation from the Code of Ethics.
        \item The authors should make sure to preserve anonymity (e.g., if there is a special consideration due to laws or regulations in their jurisdiction).
    \end{itemize}

\item {\bf Broader impacts}
    \item[] Question: Does the paper discuss both potential positive societal impacts and negative societal impacts of the work performed?
    \item[] Answer: \answerNA{}
    \item[] Justification: This work is foundational algorithmic research on HTN search heuristics and has no immediate direct societal impact within the scope of this paper.
    \item[] Guidelines:
    \begin{itemize}
        \item The answer \answerNA{} means that there is no societal impact of the work performed.
        \item If the authors answer \answerNA{} or \answerNo, they should explain why their work has no societal impact or why the paper does not address societal impact.
        \item Examples of negative societal impacts include potential malicious or unintended uses (e.g., disinformation, generating fake profiles, surveillance), fairness considerations (e.g., deployment of technologies that could make decisions that unfairly impact specific groups), privacy considerations, and security considerations.
        \item The conference expects that many papers will be foundational research and not tied to particular applications, let alone deployments. However, if there is a direct path to any negative applications, the authors should point it out. For example, it is legitimate to point out that an improvement in the quality of generative models could be used to generate Deepfakes for disinformation. On the other hand, it is not needed to point out that a generic algorithm for optimizing neural networks could enable people to train models that generate Deepfakes faster.
        \item The authors should consider possible harms that could arise when the technology is being used as intended and functioning correctly, harms that could arise when the technology is being used as intended but gives incorrect results, and harms following from (intentional or unintentional) misuse of the technology.
        \item If there are negative societal impacts, the authors could also discuss possible mitigation strategies (e.g., gated release of models, providing defenses in addition to attacks, mechanisms for monitoring misuse, mechanisms to monitor how a system learns from feedback over time, improving the efficiency and accessibility of ML).
    \end{itemize}
    
\item {\bf Safeguards}
    \item[] Question: Does the paper describe safeguards that have been put in place for responsible release of data or models that have a high risk for misuse (e.g., pre-trained language models, image generators, or scraped datasets)?
    \item[] Answer: \answerNA{}
    \item[] Justification: The paper does not release pre-trained models, scraped datasets, or other assets with high misuse risk. The generated heuristics are domain-specific Python functions for planning search.
    \item[] Guidelines:
    \begin{itemize}
        \item The answer \answerNA{} means that the paper poses no such risks.
        \item Released models that have a high risk for misuse or dual-use should be released with necessary safeguards to allow for controlled use of the model, for example by requiring that users adhere to usage guidelines or restrictions to access the model or implementing safety filters. 
        \item Datasets that have been scraped from the Internet could pose safety risks. The authors should describe how they avoided releasing unsafe images.
        \item We recognize that providing effective safeguards is challenging, and many papers do not require this, but we encourage authors to take this into account and make a best faith effort.
    \end{itemize}

\item {\bf Licenses for existing assets}
    \item[] Question: Are the creators or original owners of assets (e.g., code, data, models), used in the paper, properly credited and are the license and terms of use explicitly mentioned and properly respected?
    \item[] Answer: \answerYes{}
    \item[] Justification: All assets are cited with their original papers: \pytrich{}~\citep{PutrichMeneguzziPereira2025}, which is based on the Pyperplan planner (GPL~3 license); \panda{}~\citep{BercherKeenBiundo2014,HoellerBercherBehnkeEtAl2020} (BSD~3 license); the IPC 2020 HTN benchmark domains~\citep{BehnkeHoellerBercher2021} and the IPC 2023 Lamps domain~\citep{TaitlerAlfordEspasaEtAl2024}. LLM APIs are accessed under their standard published terms of use.
    \item[] Guidelines:
    \begin{itemize}
        \item The answer \answerNA{} means that the paper does not use existing assets.
        \item The authors should cite the original paper that produced the code package or dataset.
        \item The authors should state which version of the asset is used and, if possible, include a URL.
        \item The name of the license (e.g., CC-BY 4.0) should be included for each asset.
        \item For scraped data from a particular source (e.g., website), the copyright and terms of service of that source should be provided.
        \item If assets are released, the license, copyright information, and terms of use in the package should be provided. For popular datasets, \url{paperswithcode.com/datasets} has curated licenses for some datasets. Their licensing guide can help determine the license of a dataset.
        \item For existing datasets that are re-packaged, both the original license and the license of the derived asset (if it has changed) should be provided.
        \item If this information is not available online, the authors are encouraged to reach out to the asset's creators.
    \end{itemize}

\item {\bf New assets}
    \item[] Question: Are new assets introduced in the paper well documented and is the documentation provided alongside the assets?
    \item[] Answer: \answerYes{}
    \item[] Justification: The generated heuristics and the result files are part of the released repository (see the code question above). The heuristic interface they implement is documented in Appendix~\ref{app:api} and the generation procedure is described in Section~\ref{sec:method}.
    \item[] Guidelines:
    \begin{itemize}
        \item The answer \answerNA{} means that the paper does not release new assets.
        \item Researchers should communicate the details of the dataset\slash code\slash model as part of their submissions via structured templates. This includes details about training, license, limitations, etc. 
        \item The paper should discuss whether and how consent was obtained from people whose asset is used.
        \item At submission time, remember to anonymize your assets (if applicable). You can either create an anonymized URL or include an anonymized zip file.
    \end{itemize}

\item {\bf Crowdsourcing and research with human subjects}
    \item[] Question: For crowdsourcing experiments and research with human subjects, does the paper include the full text of instructions given to participants and screenshots, if applicable, as well as details about compensation (if any)?
    \item[] Answer: \answerNA{}
    \item[] Justification: The paper involves no crowdsourcing or research with human subjects; all experiments are automated planner evaluations on benchmark problems.
    \item[] Guidelines:
    \begin{itemize}
        \item The answer \answerNA{} means that the paper does not involve crowdsourcing nor research with human subjects.
        \item Including this information in the supplemental material is fine, but if the main contribution of the paper involves human subjects, then as much detail as possible should be included in the main paper. 
        \item According to the NeurIPS Code of Ethics, workers involved in data collection, curation, or other labor should be paid at least the minimum wage in the country of the data collector. 
    \end{itemize}

\item {\bf Institutional review board (IRB) approvals or equivalent for research with human subjects}
    \item[] Question: Does the paper describe potential risks incurred by study participants, whether such risks were disclosed to the subjects, and whether Institutional Review Board (IRB) approvals (or an equivalent approval/review based on the requirements of your country or institution) were obtained?
    \item[] Answer: \answerNA{}
    \item[] Justification: The paper involves no human subjects research and requires no IRB approval or equivalent review.
    \item[] Guidelines:
    \begin{itemize}
        \item The answer \answerNA{} means that the paper does not involve crowdsourcing nor research with human subjects.
        \item Depending on the country in which research is conducted, IRB approval (or equivalent) may be required for any human subjects research. If you obtained IRB approval, you should clearly state this in the paper. 
        \item We recognize that the procedures for this may vary significantly between institutions and locations, and we expect authors to adhere to the NeurIPS Code of Ethics and the guidelines for their institution. 
        \item For initial submissions, do not include any information that would break anonymity (if applicable), such as the institution conducting the review.
    \end{itemize}

\item {\bf Declaration of LLM usage}
    \item[] Question: Does the paper describe the usage of LLMs if it is an important, original, or non-standard component of the core methods in this research? Note that if the LLM is used only for writing, editing, or formatting purposes and does \emph{not} impact the core methodology, scientific rigor, or originality of the research, declaration is not required.
    \item[] Answer: \answerYes{}
    \item[] Justification: LLMs are the core method of this work. Section~\ref{sec:method} describes how LLMs are prompted to generate Python heuristic functions, Section~\ref{sec:setup} lists the nine models evaluated and their providers, and Appendix~\ref{app:expdetails} gives the generation settings and the candidate counts per model and domain.
    \item[] Guidelines:
    \begin{itemize}
        \item The answer \answerNA{} means that the core method development in this research does not involve LLMs as any important, original, or non-standard components.
        \item Please refer to our LLM policy in the NeurIPS handbook for what should or should not be described.
    \end{itemize}

\end{enumerate}

\end{document}